%% file: tedigan.tex
\documentclass[10pt,journal,compsoc]{IEEEtran}
\ifCLASSOPTIONcompsoc
  \usepackage[nocompress]{cite}
\else
  \usepackage{cite}
\fi
\usepackage{amsmath}
\usepackage{amssymb}
\usepackage{amsthm}
\usepackage{graphicx} 
\usepackage{booktabs}
\usepackage{url}
\usepackage{xspace}
\usepackage{xcolor}
\usepackage{multirow}

\usepackage[pagebackref=false,breaklinks=true,colorlinks=true,citecolor=blue,bookmarks=false]{hyperref}

\usepackage{pifont}

\input{newcommand}
\input{figure}

\begin{document}

\input{sections/authors}
\input{sections/abstract}

\maketitle
\IEEEdisplaynontitleabstractindextext
\IEEEpeerreviewmaketitle

\input{sections/introduction}
\input{sections/related-work}
\input{sections/method-A}
\input{sections/method-B}
\input{sections/experiment}
\input{sections/discussion}
\input{sections/conclusion}

\bibliographystyle{IEEEtran}
\bibliography{IEEEfull}

\ifCLASSOPTIONcaptionsoff
 \newpage
\fi

\end{document}

%% file: newcommand.tex
\renewcommand\thetable{\Roman{table}} 

\newcommand{\w}{{\rm\bf w}}     
\newcommand{\W}{\mathcal{W}}    
  
\newcommand{\x}{{\rm\bf x}}    
 
\newcommand{\X}{\mathcal{X}}

\newcommand{\Loss}{\mathcal{L}} 

\newcommand{\E}{\mathbb{E}}     
\newcommand{\z}{{\rm\bf z}}     
\newcommand{\Z}{\mathcal{Z}}    
\def\eg{\emph{e.g.}}
\def\ie{\emph{i.e.}}
\def\etal{\emph{et al.}}
\def\etc{\emph{etc}}

%% file: figure.tex
\newcommand{\figteaser}{
\begin{figure}[t]
\centering
\includegraphics[width=1.0\linewidth]{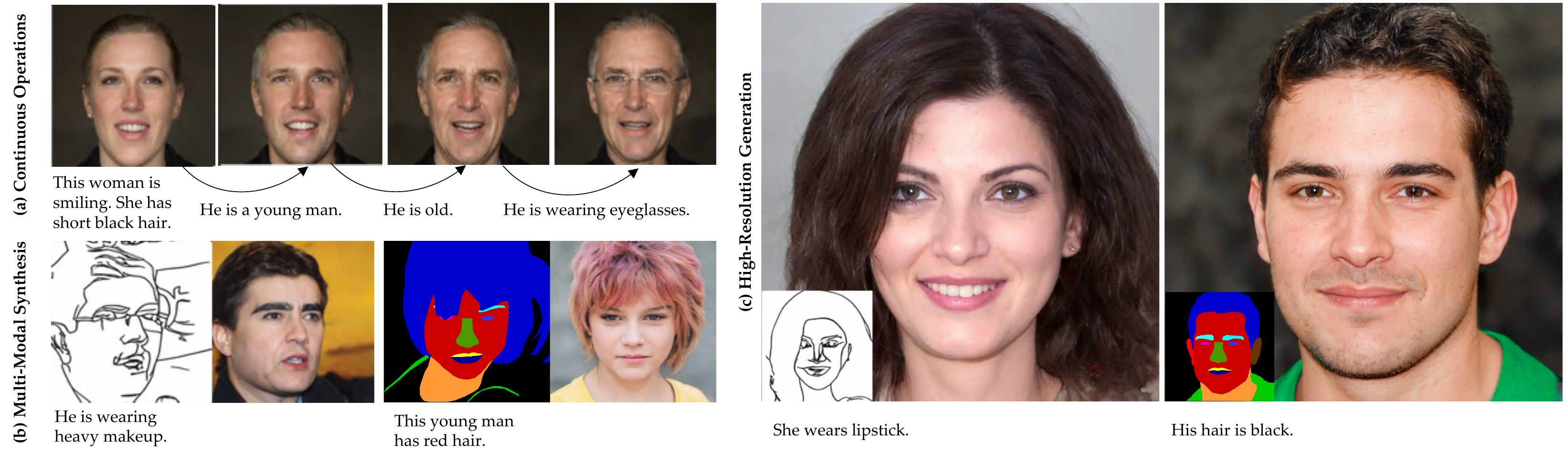}
\caption{Our TediGAN unifies text-guided image generation and manipulation into one framework, leading to continuous operations from generation to manipulation (a), inherent support of image synthesis from multi-modal inputs (b), 
and high-resolution synthesis (c).
}
\label{fig:teaser}
\end{figure}
}

\newcommand{\figoverview}{
\begin{figure*}[t]
\centering
\includegraphics[width=1.0\linewidth]{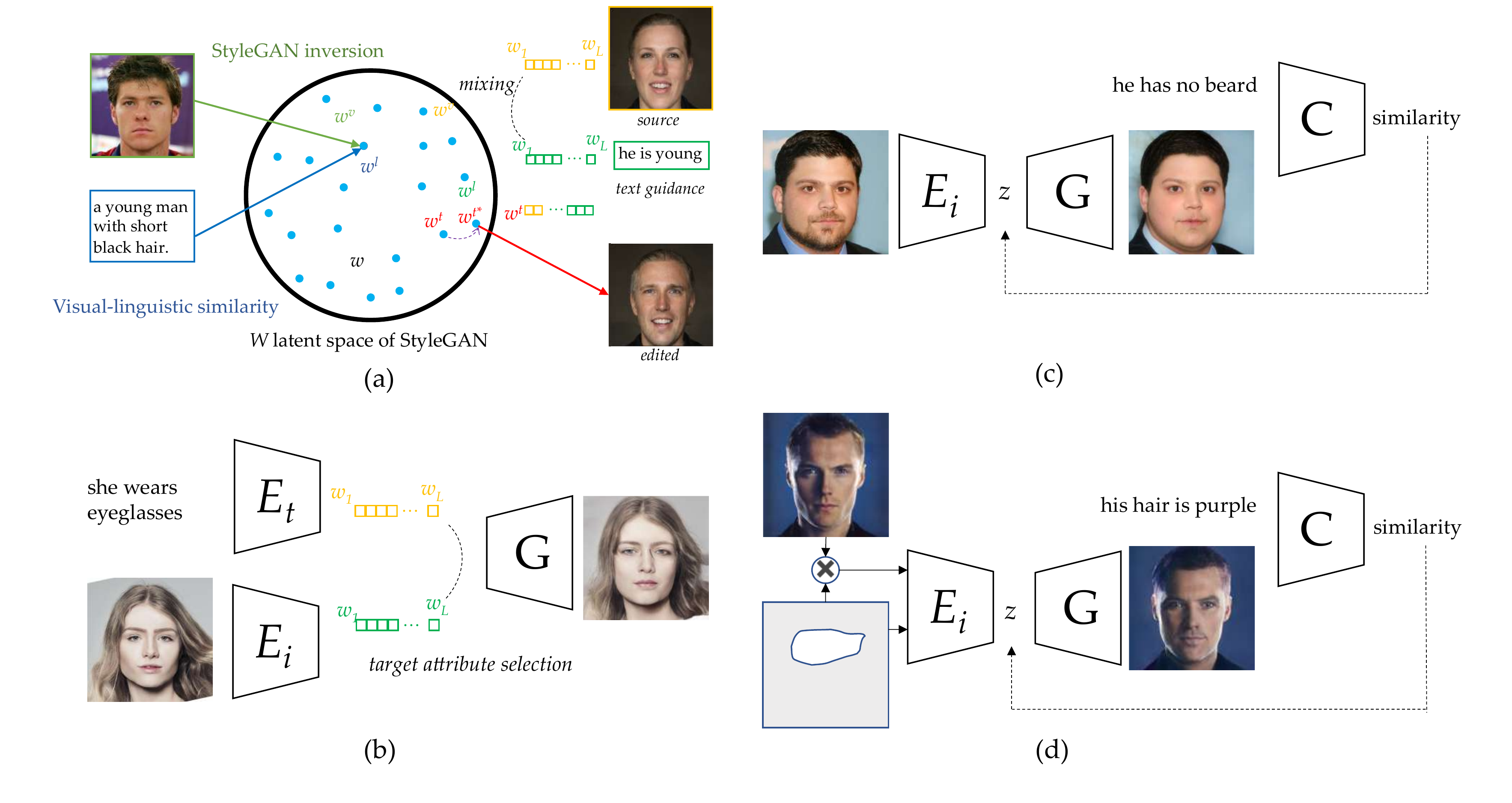}
\caption{Overview of our proposed method. We propose two strategies to use a pretrained GAN model. (a) demonstrates the first strategy. The key idea is to project multi-modal embedding into the common $\mathcal{W}$ space of StyleGAN. 
Taking visual and linguistic embedding for example, with the learned the \textcolor{teal}{inversion module}, we can then learn the \textcolor{cyan}{visual-linguistic similarity}, where the visual embedding $\w^v$ and linguistic embedding $\w^l$ are expected to be close enough. 
The \textcolor{purple}{instance-level optimization} if for identity preservation. The edited image can be generated from the StyleGAN generator.
(b) illustrates the inference of text-guided image manipulation using the text encoder. Given a source image and a text guidance, we first get their embedding $\w^v$ and $\w^l$ in $\mathcal{W}$ space through corresponding encoders. We then perform style mixing for target layers and get the target latent code $\w^t$. The final $\w^{t*}$ is obtained through instance-level optimization. For image generation, we can directly obtain the results by feeding the latent codes from the text encoder into the generator.
(c) is the illustration of text-guided image manipulation using a pretrained language model.
In (d), we show that such optimization can be easily extended to support region-of-interest manipulation.
}
\label{fig:overview}
\end{figure*}
}

\newcommand{\figcompgen}{
\begin{figure*}[ht]
\centering
\includegraphics[width=1.0\linewidth]{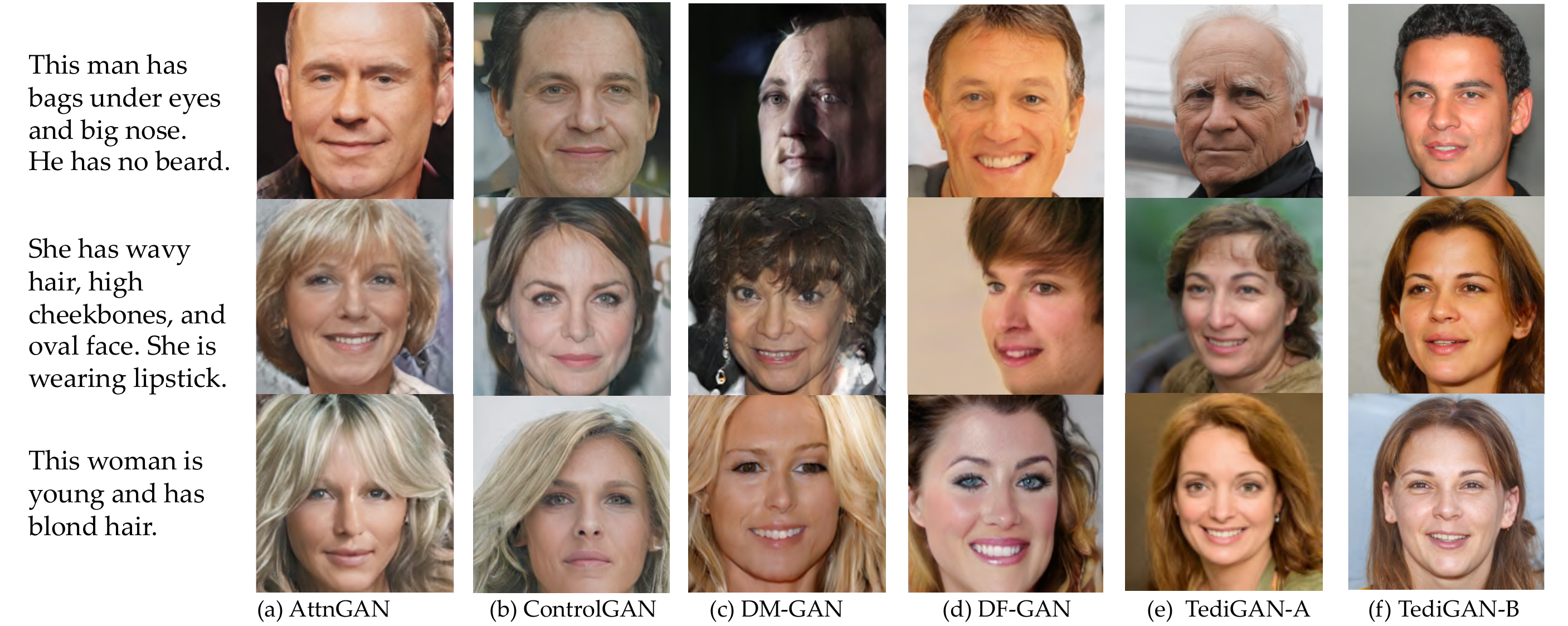}
\caption{Comparison of text-to-image generation on our Multi-modal CelebA-HQ dataset.
\textit{TediGAN-A} and \textit{-B} represents two strategies proposed in Section~\ref{subsec:train-text-encoder} and Section~\ref{subsec:pretrained-text-encoder}.
}
\label{fig:comp_gen}
\end{figure*}
}

\newcommand{\figcompman}{
\begin{figure*}[ht]
\centering
\includegraphics[width=0.95\linewidth]{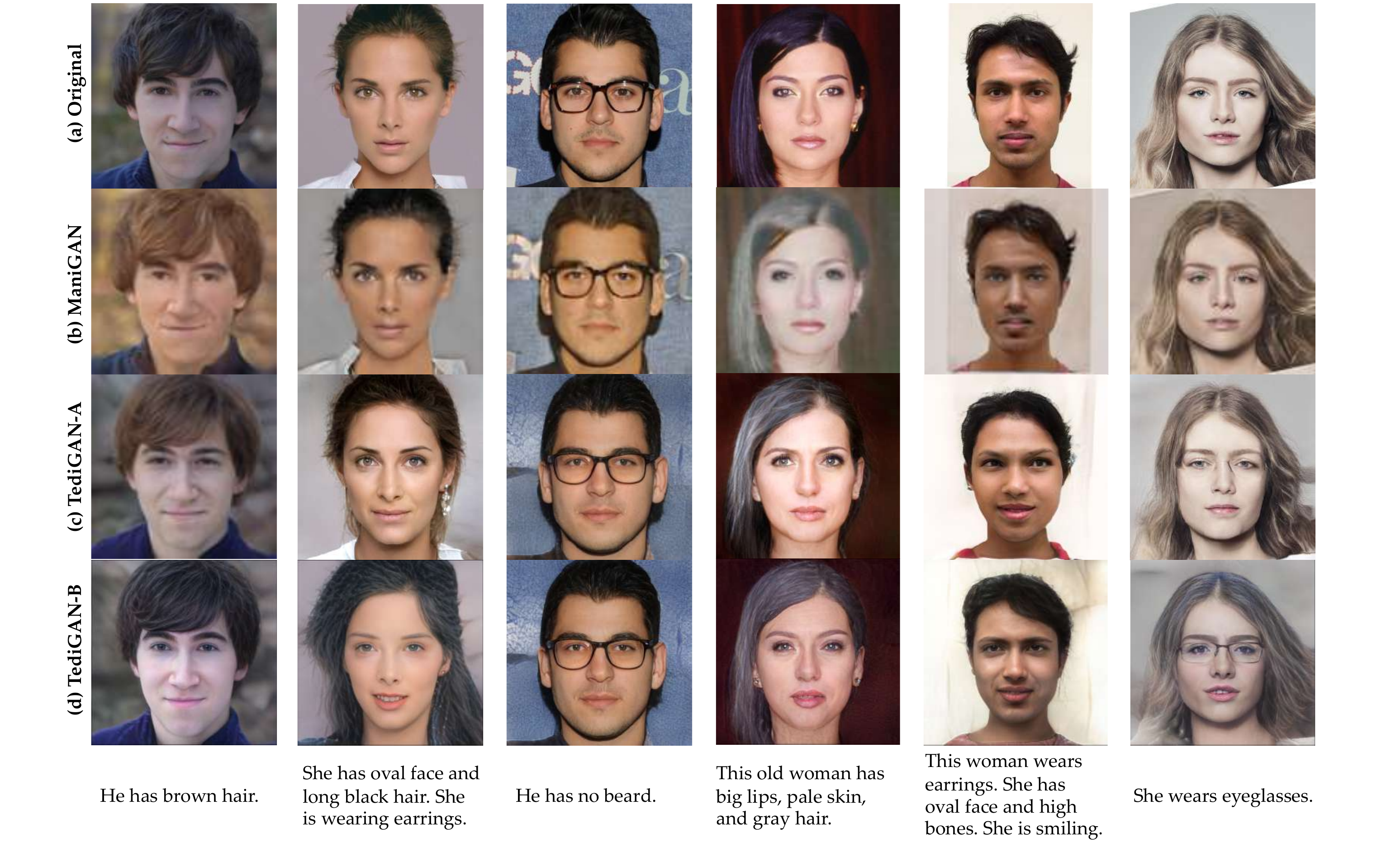}
\caption{Qualitative comparison of image manipulation using natural language descriptions.}
\label{fig:comp_man}
\end{figure*}
}

\newcommand{\figcontrol}{
\begin{figure}[t]
\centering
\includegraphics[width=1.0\linewidth]{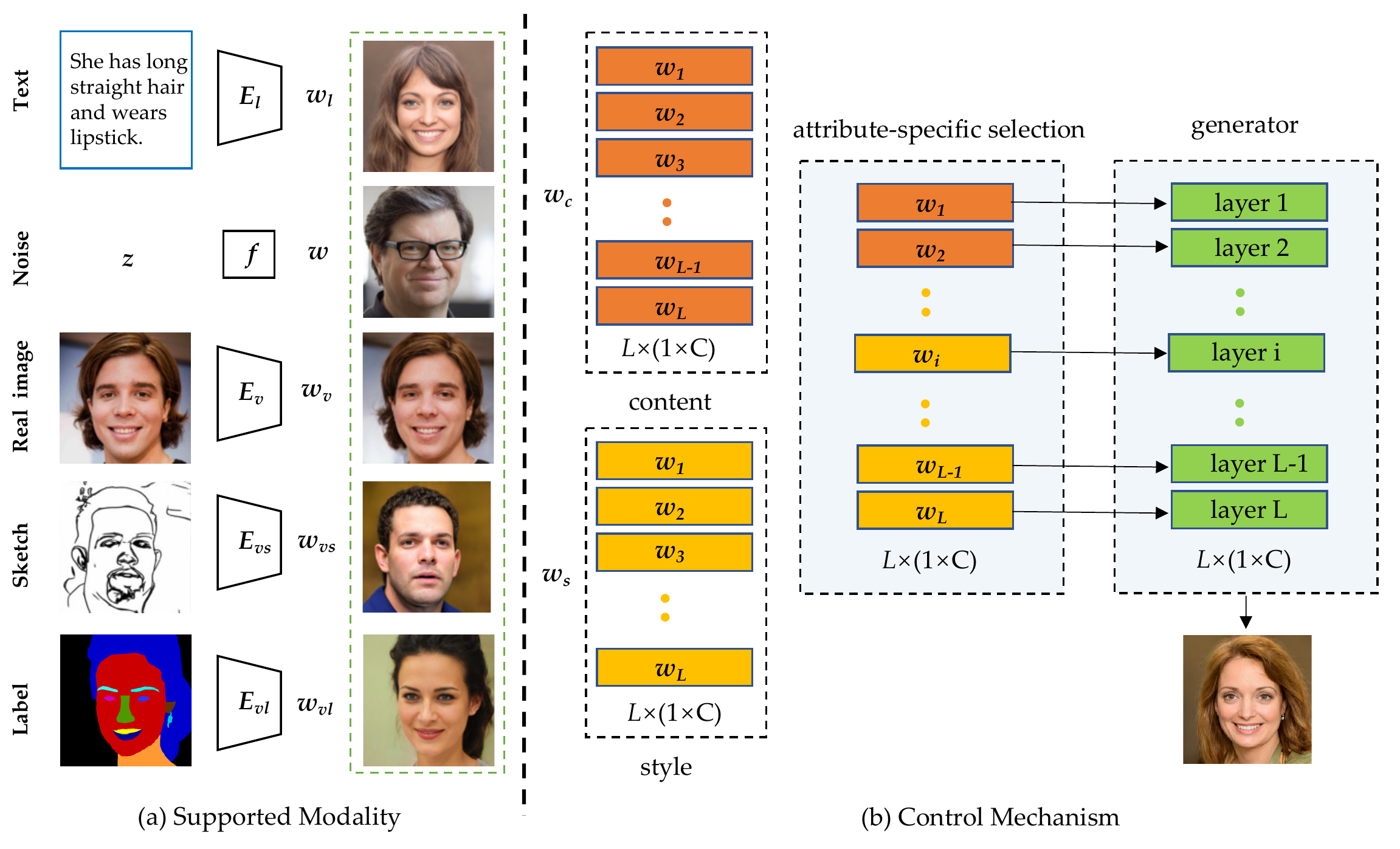}
\caption{Control mechanism of our TediGAN framework.
Different layer in the StyleGAN generator represents different attributes. 
Changing the value of a certain layer would change the corresponding attributes of the image.
Since the texts and images are mapped into the common latent space, we can synthesize images with certain attributes by selecting attribute-specific layers.
The control mechanism mixes layers of the style code $\w^s$ by partially replacing corresponding layers of the content $\w^c$.
When $\w^s$ is a randomly sampled latent code, it is the text-to-image generation and when $\w^s$ is the image embedding, it performs text-guided image manipulation. 
}
\label{fig:control_mechanism}
\end{figure}
}

\newcommand{\figlayer}{
\begin{figure*}[t]
\includegraphics[width=0.95\linewidth]{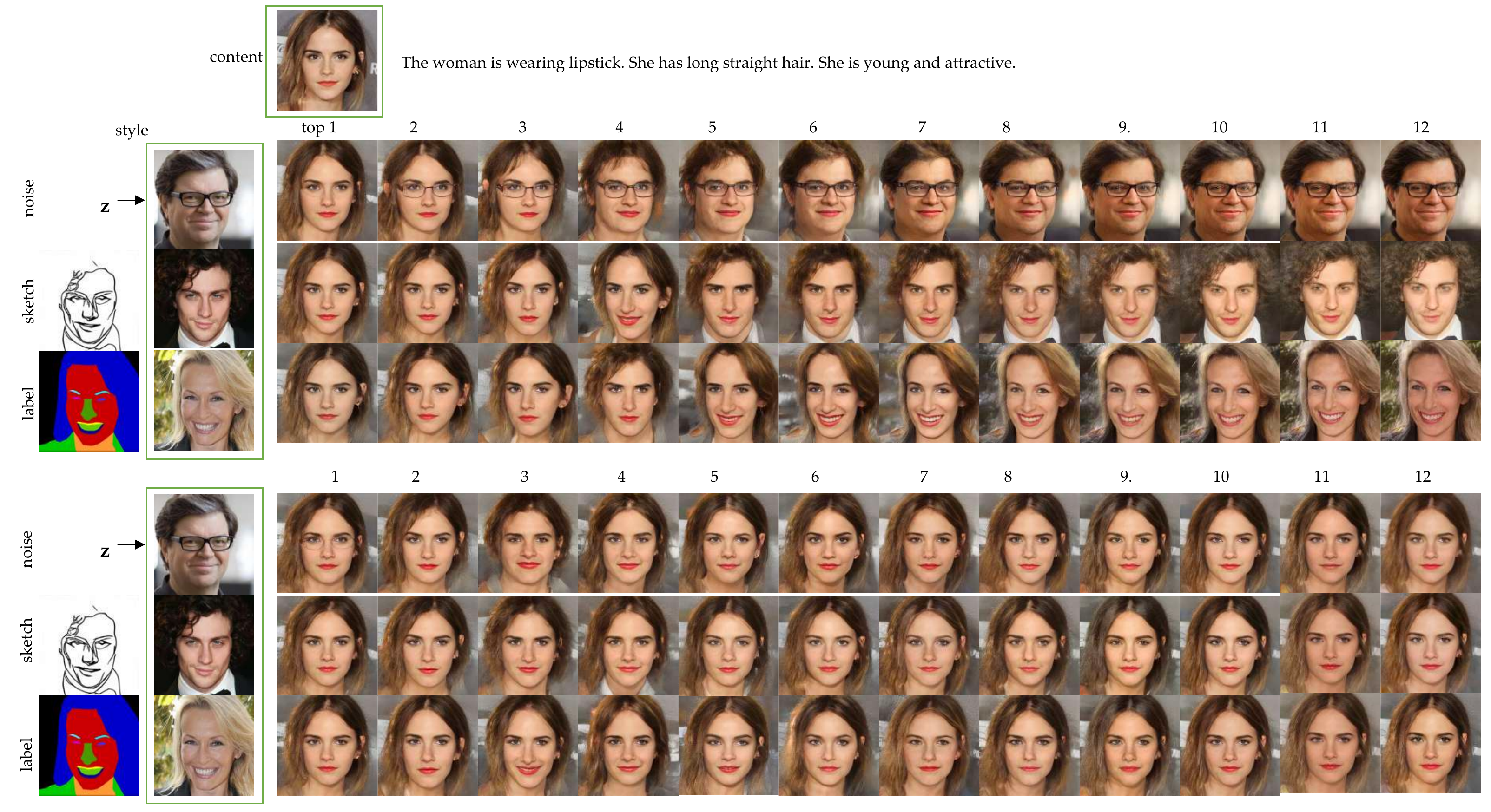}
\centering
\caption{Layerwise analysis of control mechanism.}
\label{fig:layerwise_analysis}
\end{figure*}
}

\newcommand{\figdiverse}{
\begin{figure*}[t]
\includegraphics[width=1.0\linewidth]{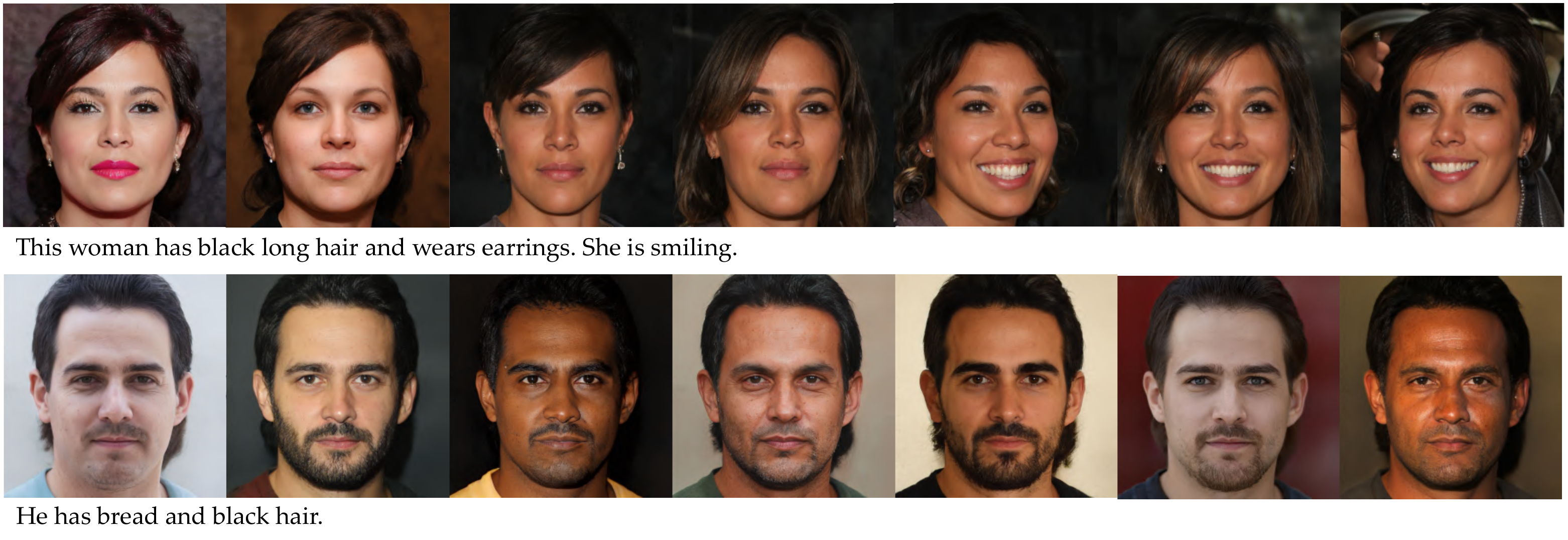}
\centering
\caption{Diverse text-to-image generation.}
\label{fig:diverse_image}
\end{figure*}
}

\newcommand{\figoodsketch}{
\begin{figure*}[t]
\centering
\includegraphics[width=1.0\linewidth]{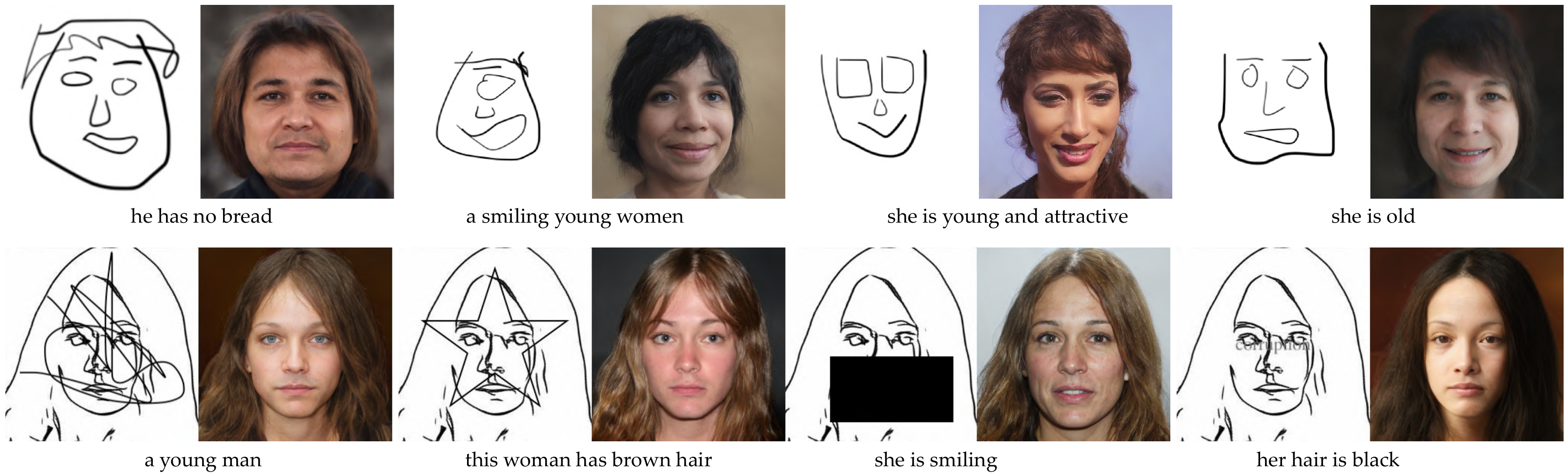}
\caption{Image synthesis from sketches of different styles.
The sketches in the first roware real human-drawn sketches.
The second are degradations of one sample randomly chosen from our sketch data, from top to bottom adding noisy scribbles, mask, colorful text, and irregular shape.}
\label{fig:ood_sketch}
\end{figure*}
}

\newcommand{\figdata}{
\begin{figure}[t]
\includegraphics[width=0.95\linewidth]{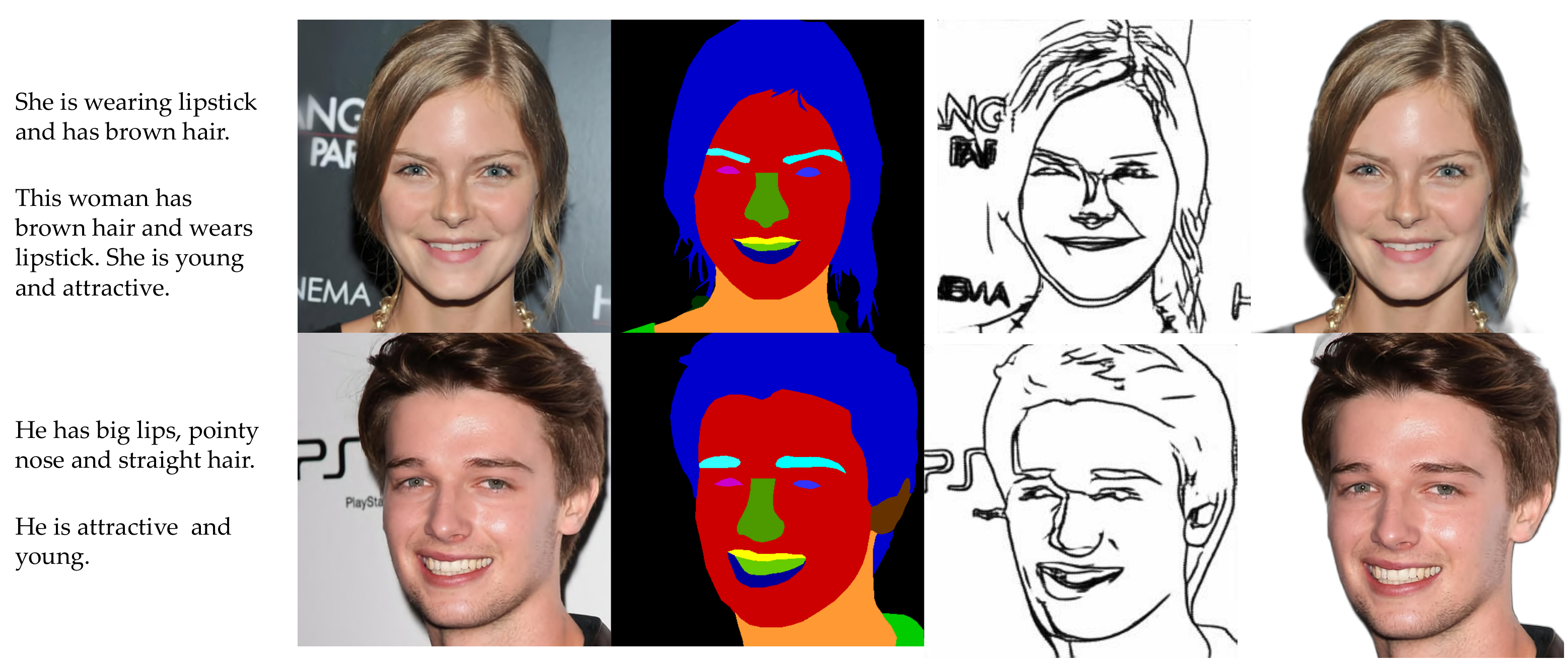}
\centering
\caption{Data sample of the Multi-Modal-CelebA-HQ dataset. From left to right are textual descriptions, real images, semantic labels, sketches, and images with transparent background.}
\label{fig:sample_data}
\end{figure}
}

\newcommand{\tablayer}{
\begin{table}[t]
\centering
\caption{The empirical layerwise analysis of a 14-layer StyleGAN generator. The 13-th and 14-th layers are omitted since there is basically no visible difference.}
\begin{tabular}{cc|cc}
\toprule
$n$-th & attribute & $n$-th & attribute \\
\midrule
1 &eye glasses & 7 & hair color  \\
2 &head pose & 8 & face color  \\
3 &face shape & 9 & age \\
4 &hair length, nose, lip & 10 & gender \\
5 &cheekbones  & 11 & micro features \\
6 &chin & 12 & micro features \\
\bottomrule
\end{tabular}
\label{tab:layerwise_analysis}
\end{table}
}

\newcommand{\tabquangen}{
\begin{table}[t]
\centering
\caption{Quantitative comparison of text-to-image generation. 
We compare the state-of-the-arts and our method in terms of FID, LPIPS, accuracy (Acc.), and realism (Real.). $\downarrow$ means the lower the better while $\uparrow$ means the opposite.}
\begin{tabular}{c|ccccc}
\toprule
Method &FID $\downarrow$ &LPIPS $\downarrow$ & Acc. (\%) $\uparrow$ &Real. (\%) $\uparrow$\\
\midrule
AttnGAN~\cite{xu2018attngan} &125.98 &0.512 &13.0 &11.9 \\
ControlGAN~\cite{li2019control} &116.32 &0.522 &14.6 &13.1 \\
DFGAN~\cite{tao2020dfgan} &137.60 &0.581 &17.3 &14.5 \\
DM-GAN~\cite{zhu2019dmgan} &131.05 &0.544 &16.4 &16.9 \\
TediGAN-A & 106.37 &\textbf{0.456} &18.4 &\textbf{22.6} \\
TediGAN-B &\textbf{101.42} & 0.461 &\textbf{20.4}&21.0 \\
\bottomrule
\end{tabular}
\label{tab:quan_gen}
\end{table}
}

\newcommand{\tabquanman}{
\begin{table}[thbp]
\centering
\caption{{Quantitative comparison of text-guided image manipulation.} We compare our method with the state-of-the-art ManiGAN~\cite{li2020manigan} in terms of FID, accuracy (Acc.), and realism (Real.).}
\begin{tabular}{c|cccc}
\hline
& Method & FID & Acc. (\%) & Real. (\%) \\ \hline
\multirow{3}{*}{CelebA} & ManiGAN~\cite{li2020manigan} &117.89 &27.4  &10.9  \\
& TediGAN-A &107.25 &34.3  &42.6  \\
& TediGAN-B &\textbf{101.27} &\textbf{38.3}  &\textbf{46.5}  \\ \hline
\multirow{3}{*}{Non-CelebA} & ManiGAN~\cite{li2020manigan} &143.39 &16.9 &7.4 \\
& TediGAN-A &135.47 &40.9  &45.2  \\
& TediGAN-B &\textbf{129.20} &\textbf{42.2}  &\textbf{47.4} \\ \hline
\multirow{3}{*}{Open-Text} & ManiGAN~\cite{li2020manigan} &141.51  &9.1 &10.9  \\
& TediGAN-A &113.57 &22.2  &43.0 \\
& TediGAN-B &\textbf{107.32} &\textbf{68.7}  &\textbf{44.8} \\ \hline
\end{tabular}
\label{tab:quan_man}
\end{table}
}

\newcommand{\figinversionresult}{
\begin{figure}[t]
\includegraphics[width=1.0\linewidth]{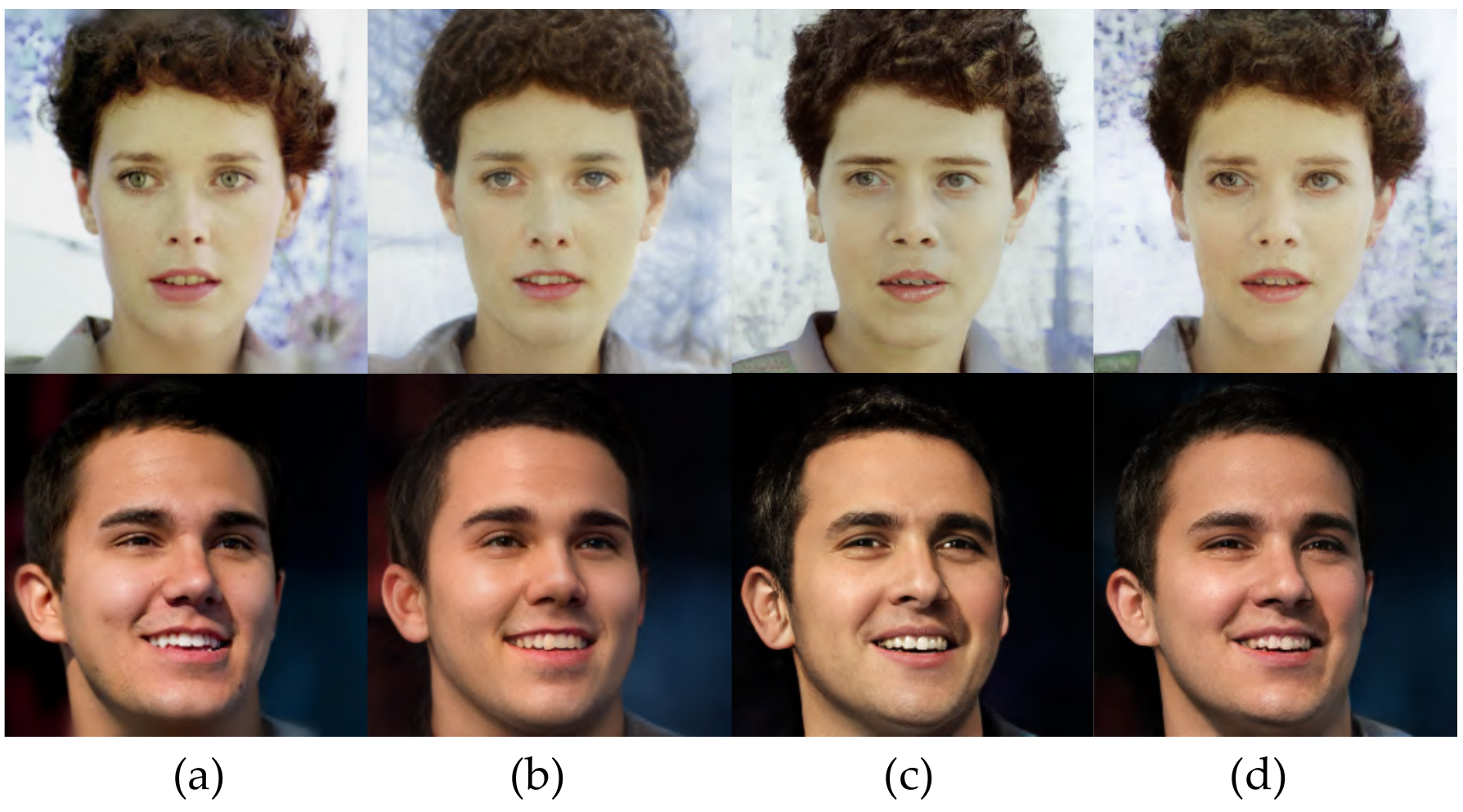}
\centering
\caption{{Inversion results.} 
(a) original image; 
(b) inversion result of pSp~\cite{richardson2020encoding}; 
(c) inversion result of our image encoder (Section~\ref{subsec:gan-inversion}); 
(d) inversion results after our optimization (Section~\ref{subsec:instance-level-optimization}).
}
\label{fig:inversion_result}
\end{figure}
}

\newcommand{\fighighres}{
\begin{figure*}[th]
\centering
\begin{minipage}[t]{0.33\textwidth}
\centering
\includegraphics[width=1.0\linewidth]{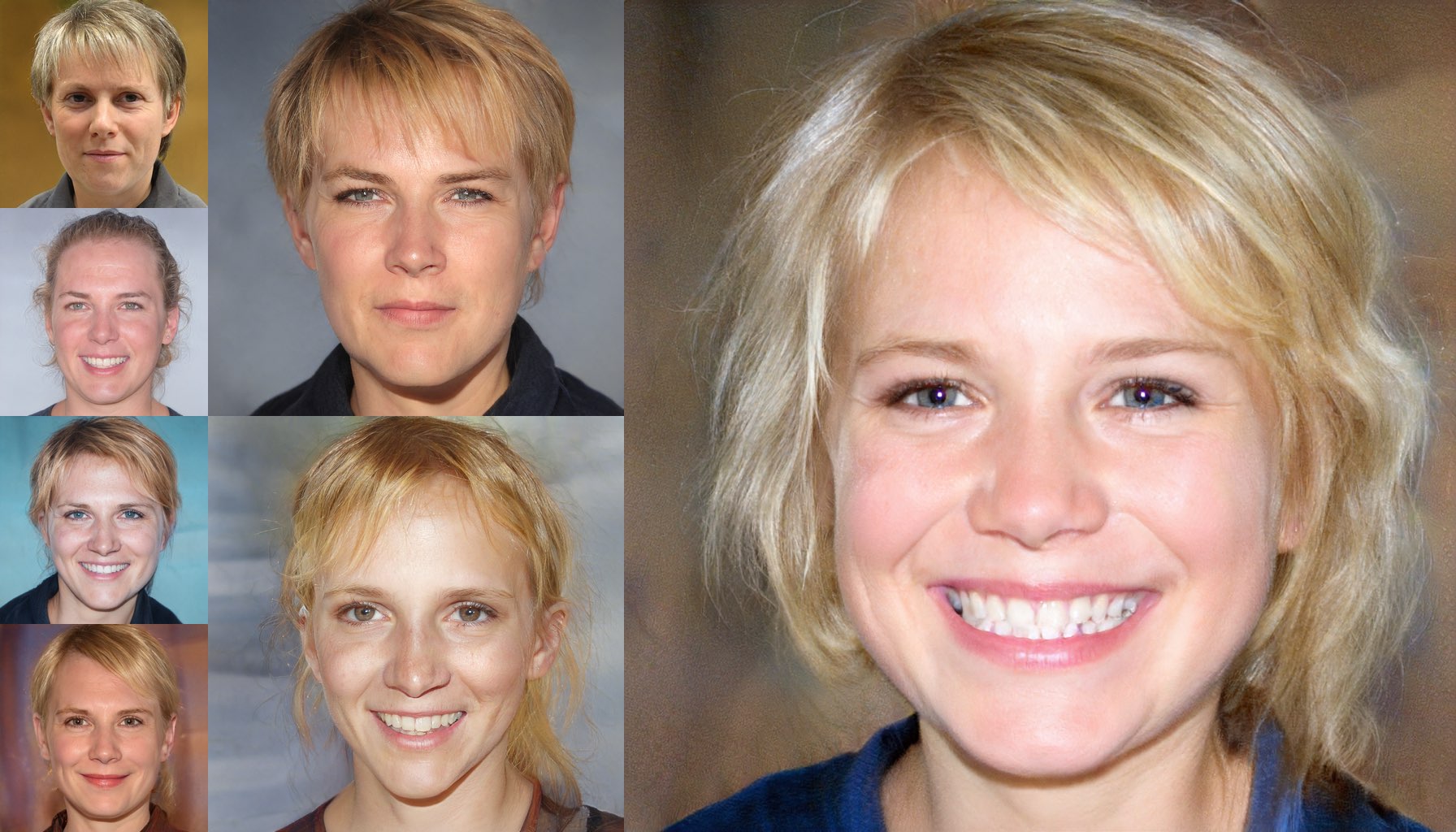}
\textit{a smiling young woman with short blonde hair}
\end{minipage}
\begin{minipage}[t]{0.33\textwidth}
\centering
\includegraphics[width=1.0\linewidth]{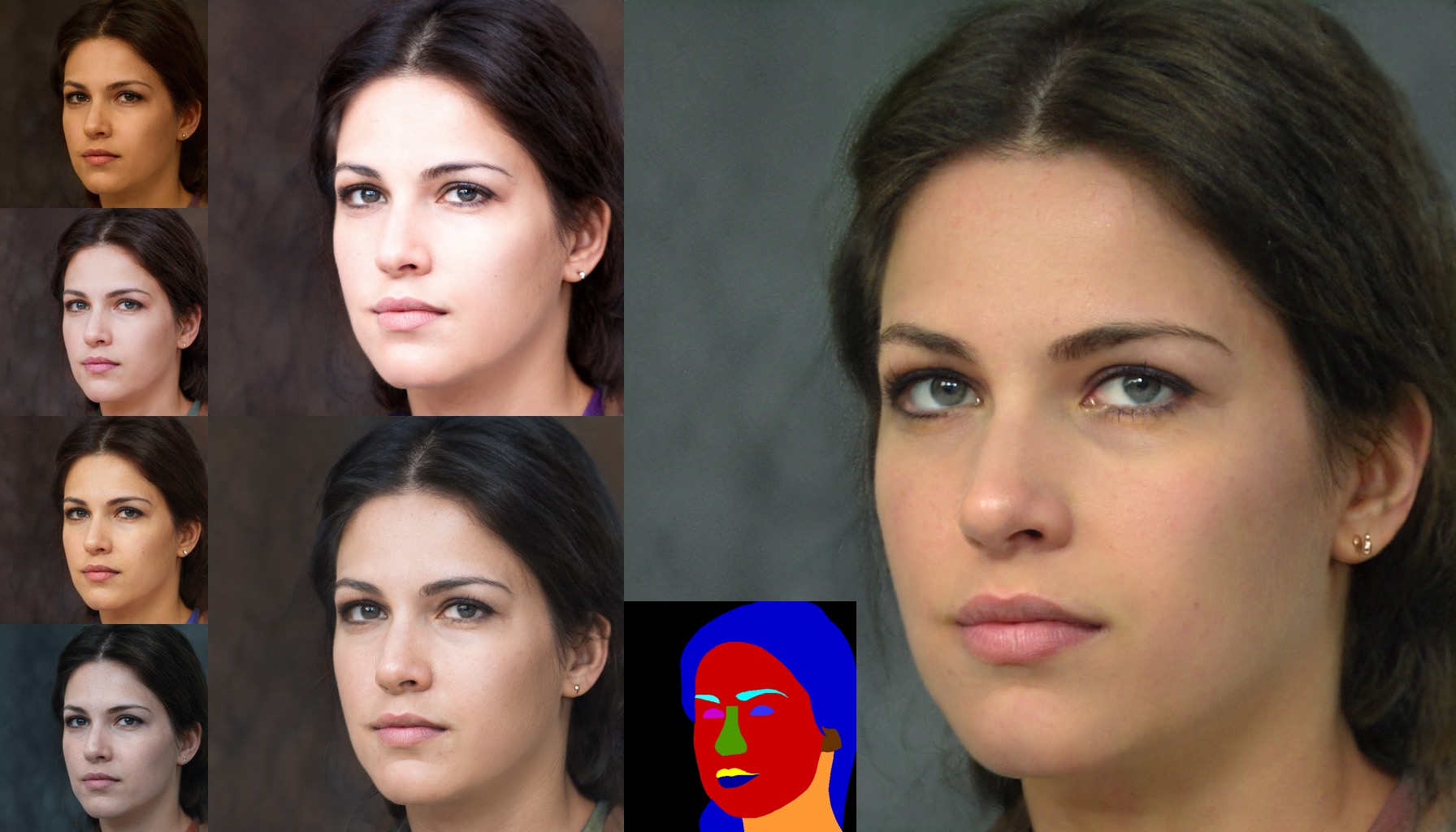}
\textit{she is young and wears earrings}
\end{minipage}
\begin{minipage}[t]{0.33\textwidth}
\centering
\includegraphics[width=1.0\linewidth]{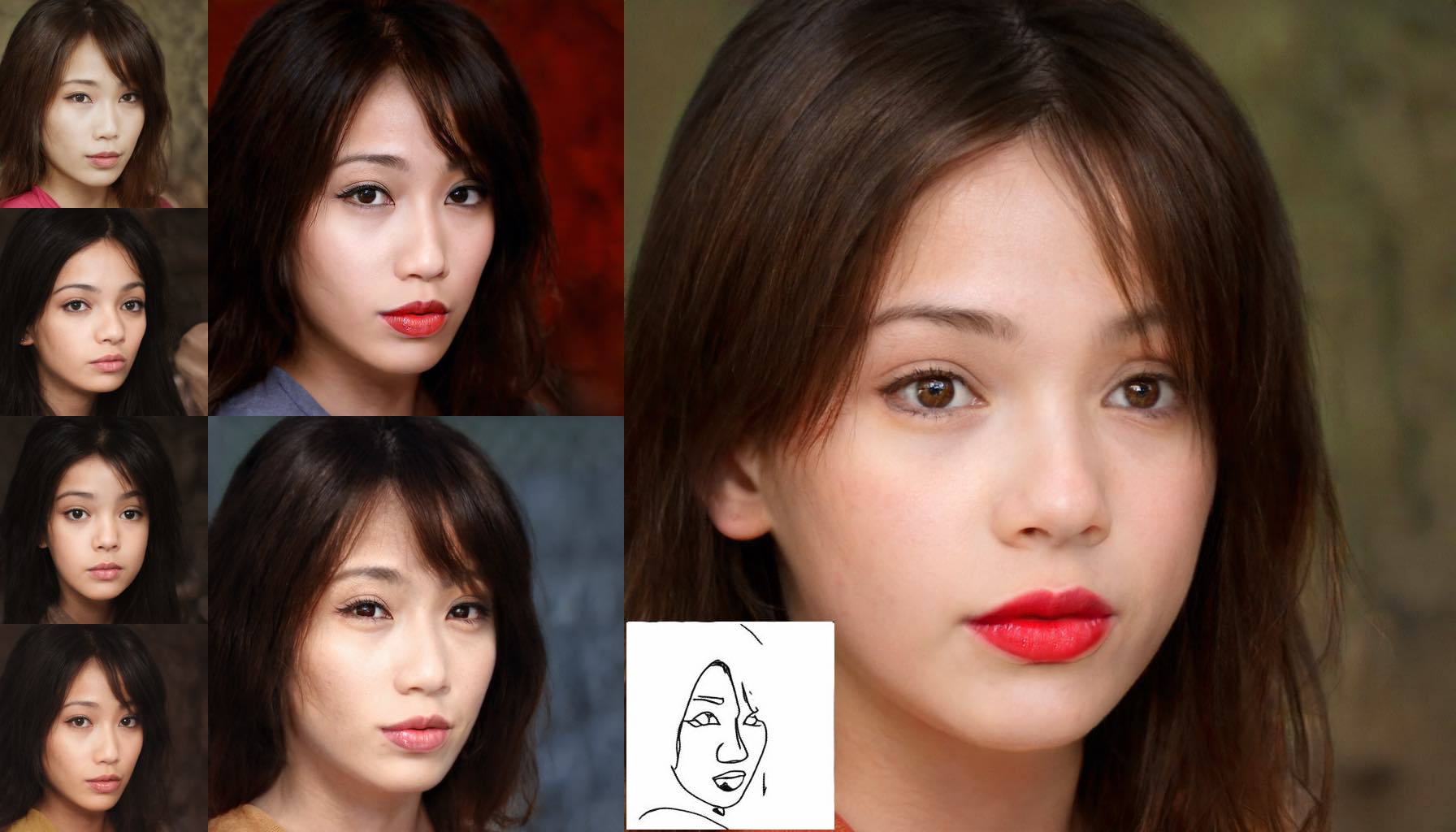}
\textit{a young woman with long black hair}
\end{minipage}
\quad
\begin{minipage}[t]{0.33\textwidth}
\centering
\includegraphics[width=1.0\linewidth]{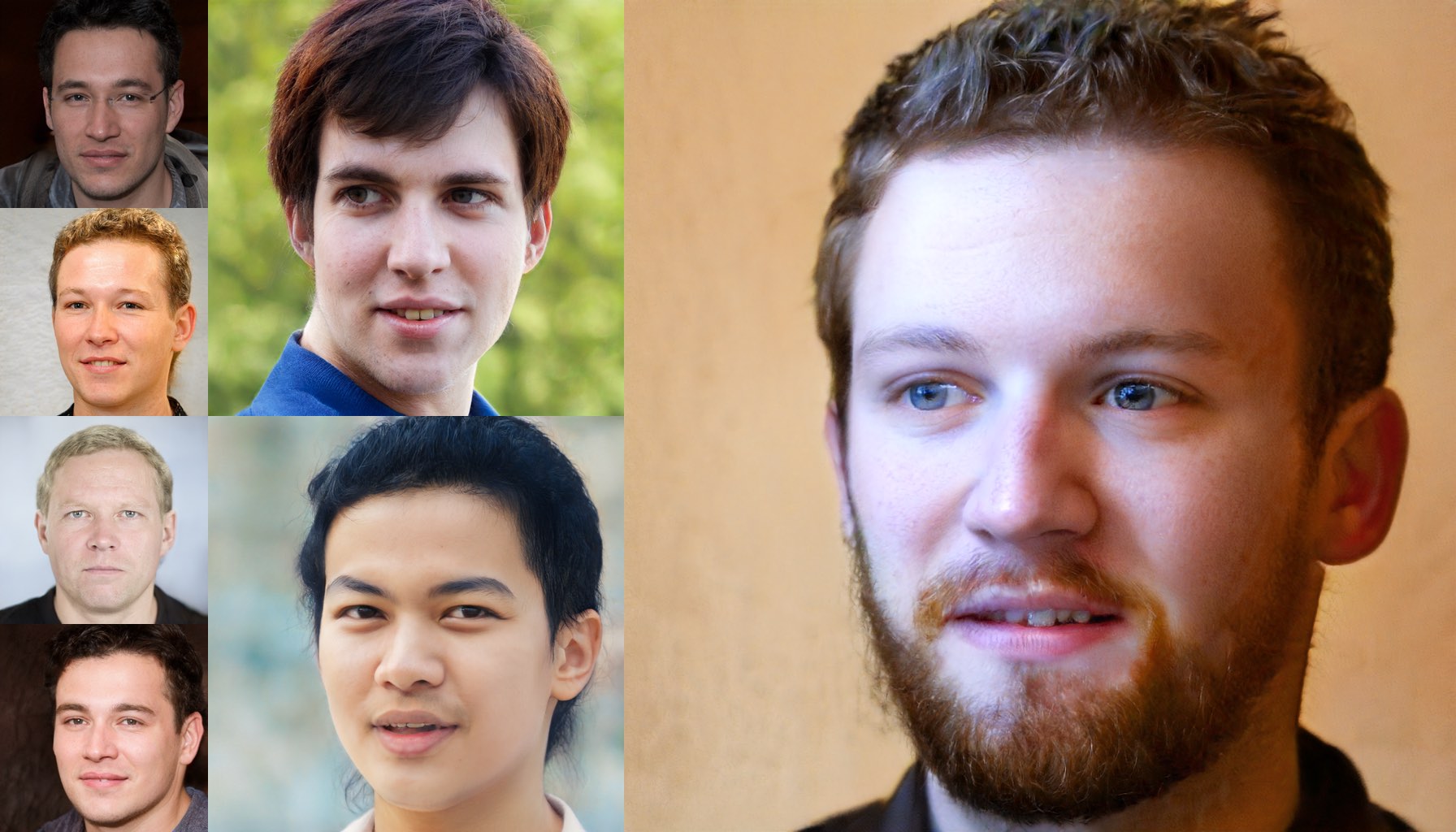}
\textit{he has high cheekbones and double chin. He is chubby}
\end{minipage}
\centering
\begin{minipage}[t]{0.33\textwidth}
\centering
\includegraphics[width=1.0\linewidth]{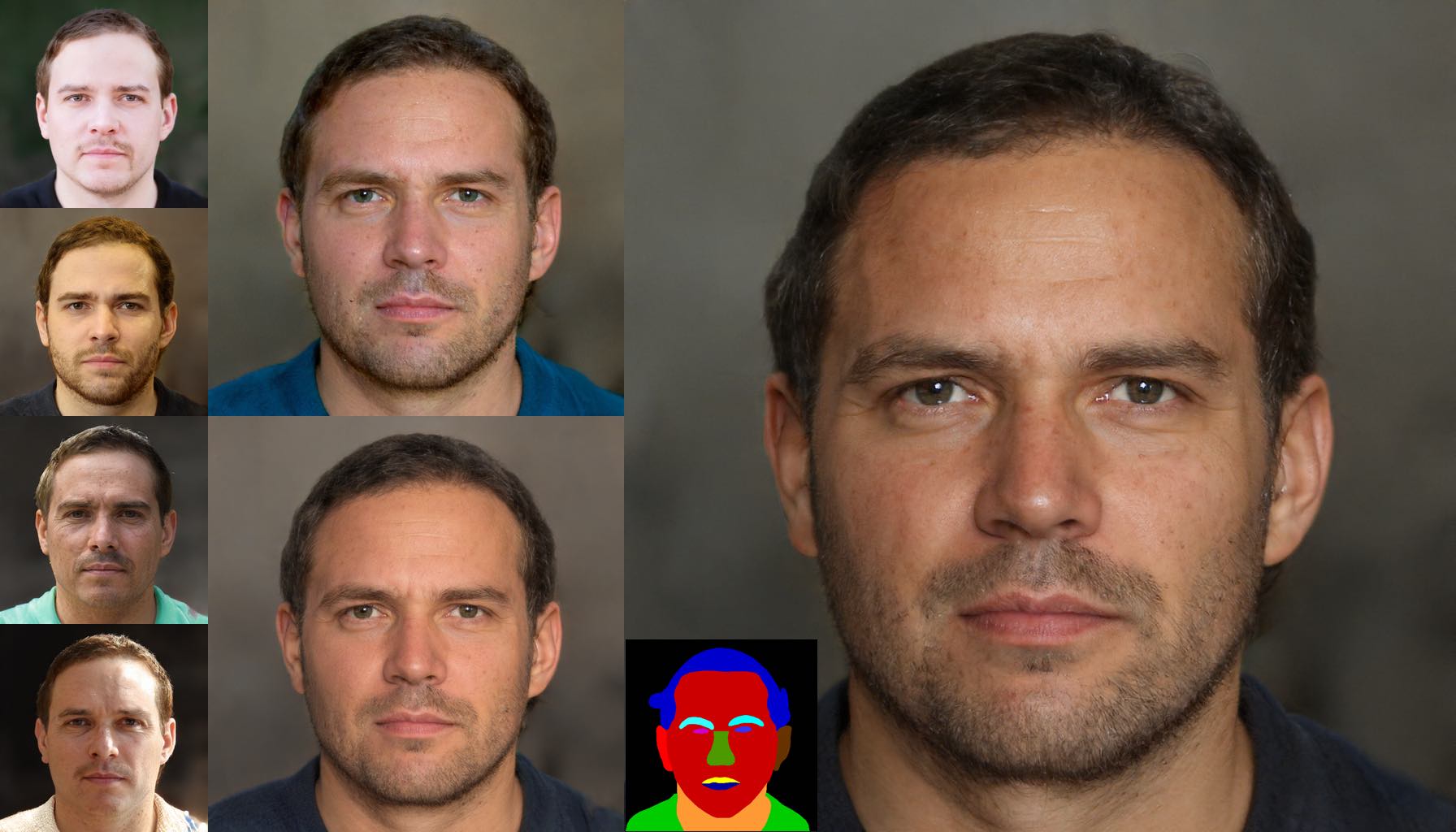}
\textit{he is young and wears beard}
\end{minipage}
\begin{minipage}[t]{0.33\textwidth}
\centering
\includegraphics[width=1.0\linewidth]{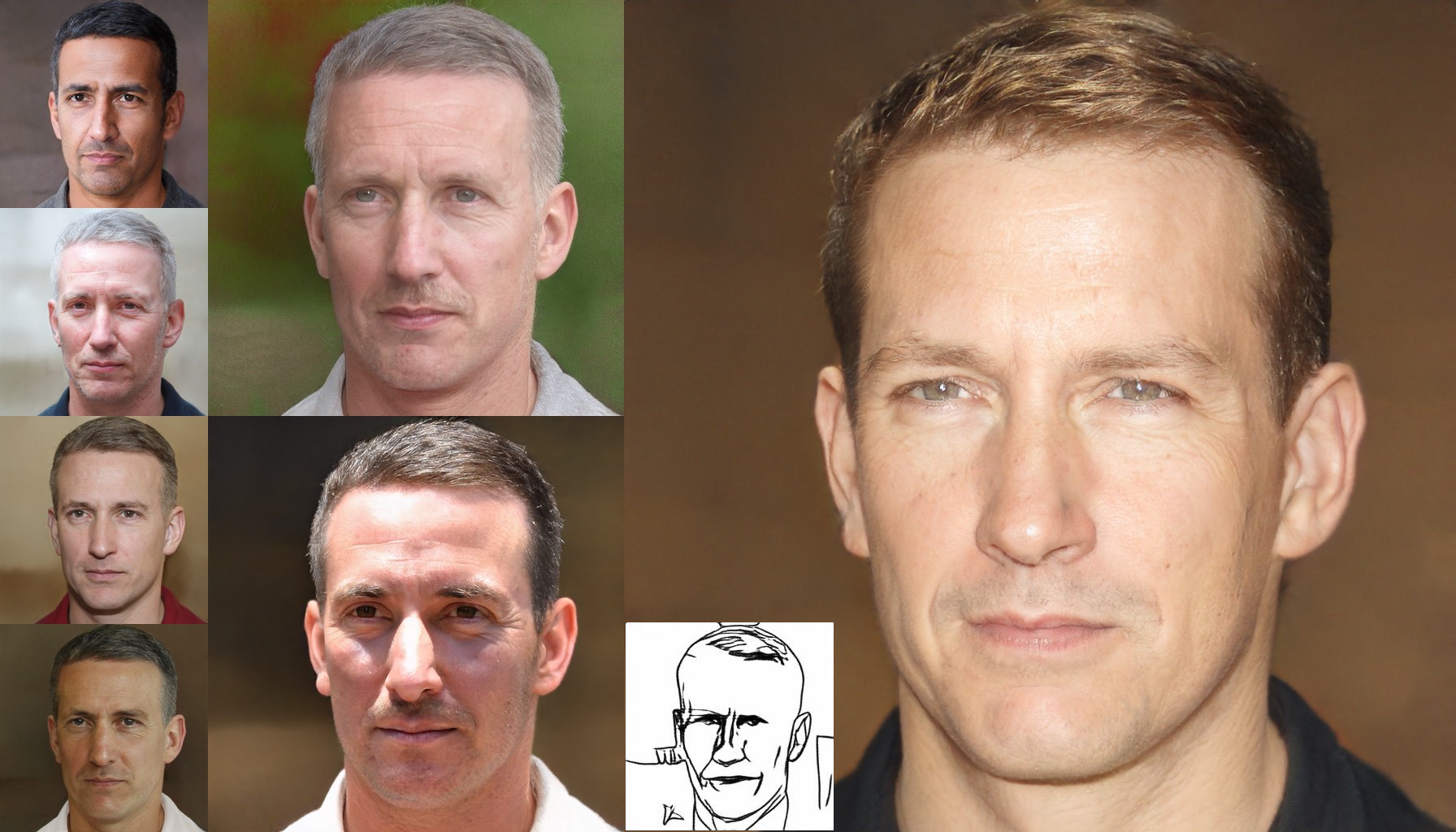}
\textit{he has a five o'clock shadow}
\end{minipage}
\caption{Diverse high-resolution results from multimodal inputs with textual guidance. Our method achieves text-guided diverse image generation and manipulation up to an unprecedented resolution at 1024 $\times$ 1024.
}
\label{fig:high-res-multi-modality}
\end{figure*}
}

\newcommand{\fignearmiss}{
\begin{figure}[t]
\begin{center}
\includegraphics[width=1.0\linewidth]{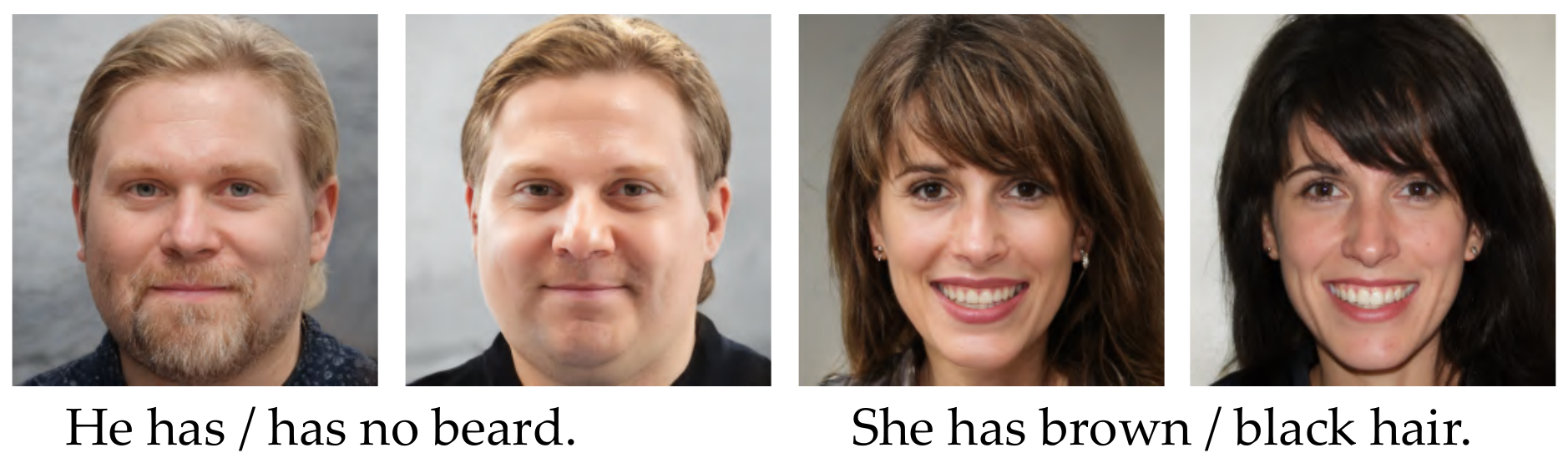}
\end{center}
\caption{Illustration of near-miss cases.}
\label{fig:near-miss}
\end{figure}
}

\newcommand{\figclip}{
\begin{figure*}[th]
\includegraphics[width=1.0\linewidth]{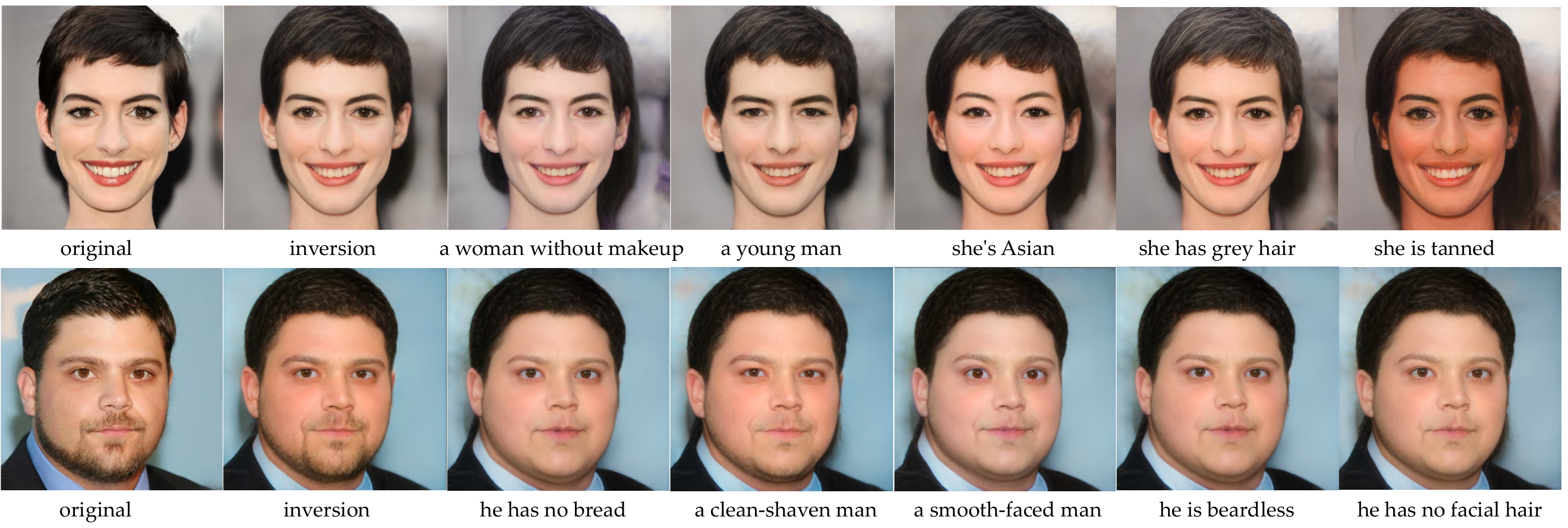}
\centering
\caption{Image manipulation results of using a pretrained text encoder CLIP~\cite{radford2021learning}.
The first row of images shows manipulated results using texts in (\textit{makeup}, \textit{young}, and \textit{grey hair}) and out (\textit{asia} and \textit{tanned}) of the proposed dataset.
The second row of images are manipulated with similar descriptions of \textit{beardless}.
}
\label{fig:with-clip}
\end{figure*}
}

\newcommand{\figgenopen}{
\begin{figure*}[t]
\centering
\includegraphics[width=1.0\linewidth]{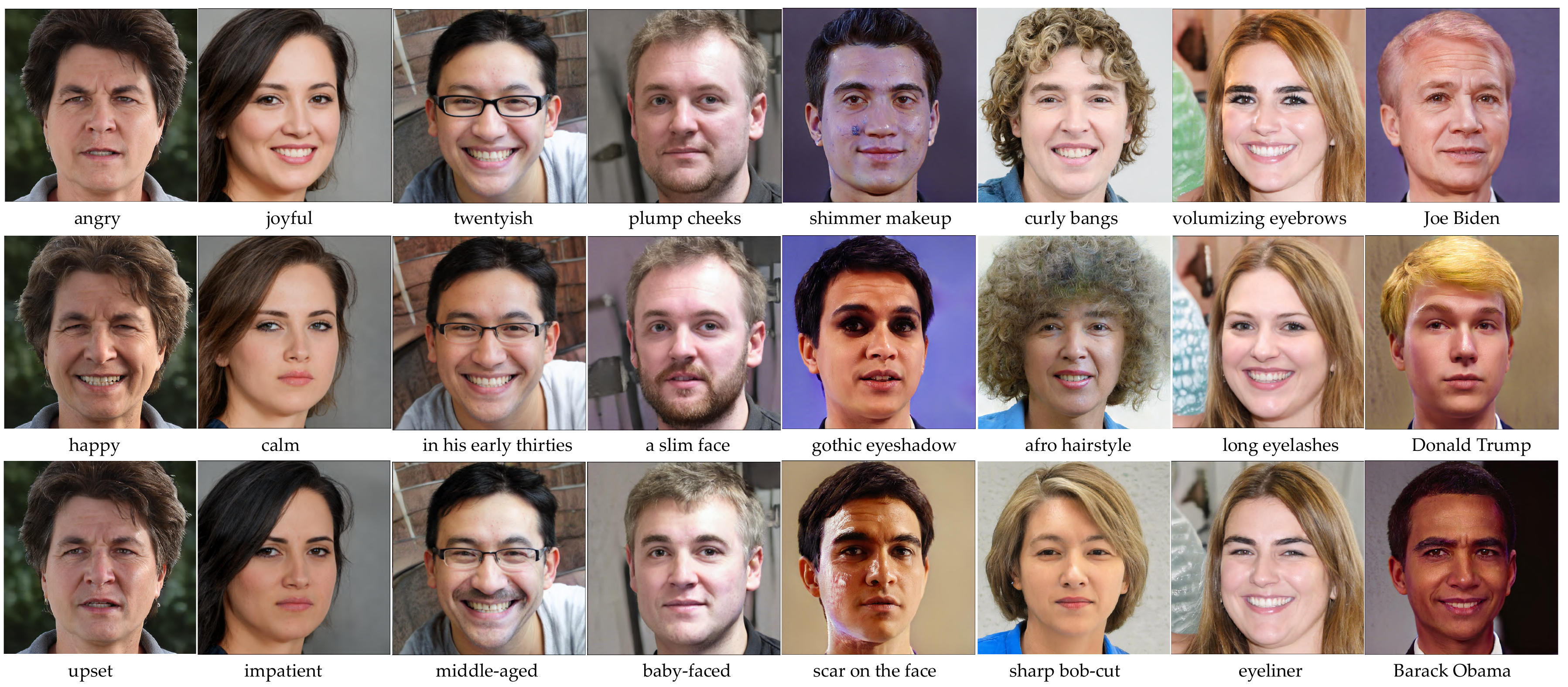}
\caption{Image generation with open-world texts.}
\label{fig:gen_open}
\end{figure*}
}

\newcommand{\figmanroi}{
\begin{figure}[t]
\centering
\includegraphics[width=1.0\linewidth]{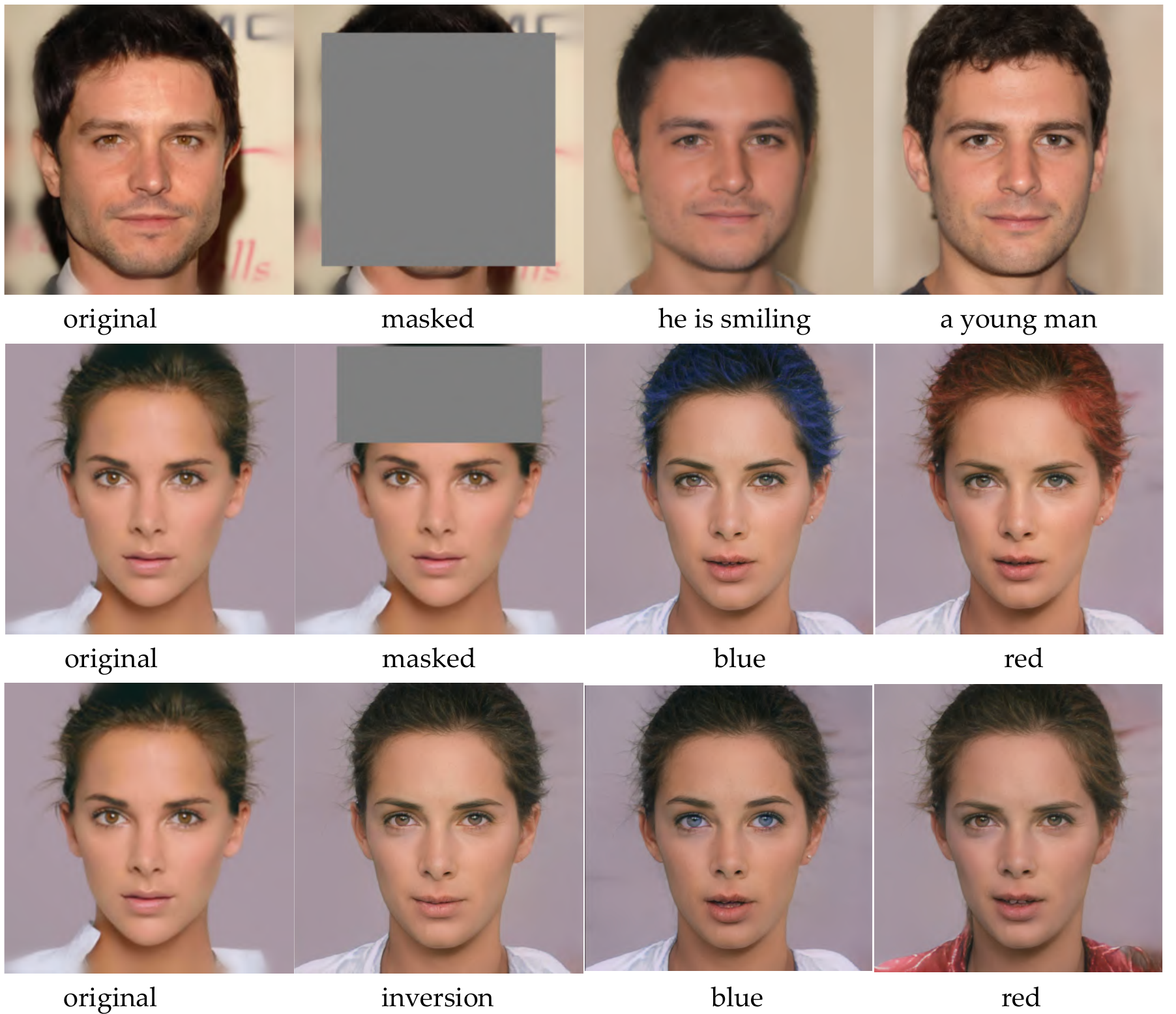}
\caption{Results of region-of-interest editing. 
The first two rows are editing results of masked regions. The third row is for comparison. For the \texttt{single-word} descriptions \textit{blue} and \textit{red}, the results in the second row are primarily restricted to the designated regions.
For the third row, since the the editing areas are not designated, the editing maybe occurs in any possible regions, \eg, blue for eyes and red for clothes.
}
\label{fig:man_roi}
\end{figure}
}

\newcommand{\figpaintclip}{
\begin{figure}[t]
\includegraphics[width=1.0\linewidth]{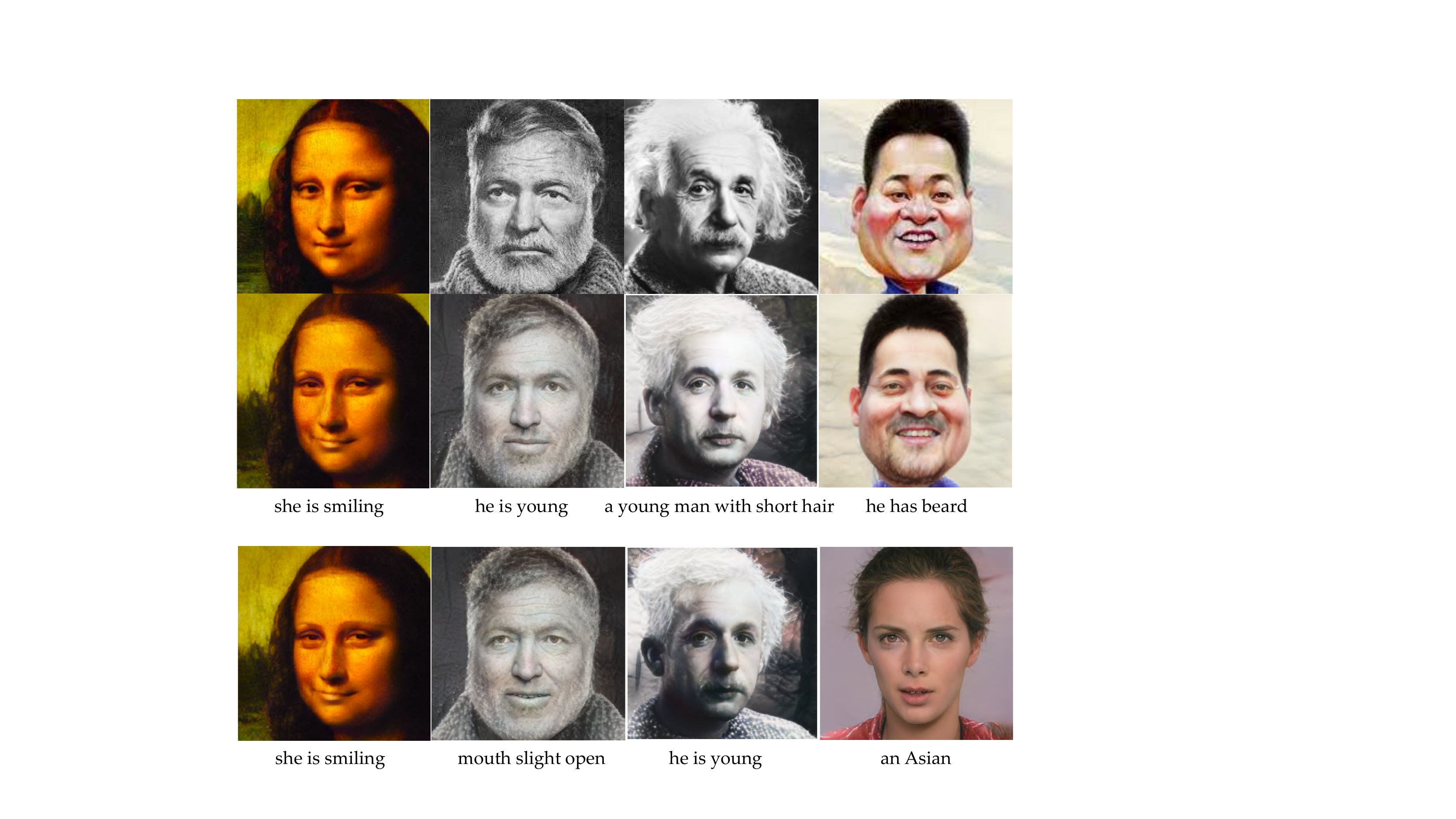}
\centering
\caption{Manipulated results for painting, caricature, and black and white photo images.
}
\label{fig:with-clip-paint}
\end{figure}
}

%% file: sections/authors.tex
\title{Towards Open-World Text-Guided Face Image Generation and Manipulation}
\author{
Weihao Xia,
Yujiu Yang*,
Jing-Hao Xue,
and Baoyuan Wu
\thanks{*Corresponding author}
\thanks{W.~Xia and Y.~Yang are with Tsinghua Shenzhen International Graduate School, Tsinghua University, China.
Email: weihaox@outlook.com, yang.yujiu@sz.tsinghua.edu.cn}
\thanks{J.-H.~Xue is with the Department of Statistical Science, University College London, UK.
Email: jinghao.xue@ucl.ac.uk}
\thanks{B.~Wu is with School of Data Science, Chinese University of Hongkong, Shenzhen, China and Secure Computing Lab of Big Data, Shenzhen Research Institute of Big Data, Shenzhen, China.
Email: wubaoyuan@cuhk.edu.cn}
}

%% file: sections/abstract.tex
\IEEEtitleabstractindextext{
\begin{abstract}
The existing text-guided image synthesis methods can only produce limited quality results with at most \mbox{$\text{256}^2$} resolution and the textual instructions are constrained in a small Corpus. 
In this work, we propose a unified framework for both face image generation and manipulation that produces diverse and high-quality images with an unprecedented resolution at~\mbox{$\text{1024}^2$} from multimodal inputs. 
More importantly, our method supports open-world scenarios, including both image and text, without any re-training, fine-tuning, or post-processing.
To be specific, we propose a brand new paradigm of text-guided image generation and manipulation based on the superior characteristics of a pretrained GAN model.
Our proposed paradigm includes two novel strategies.
The first strategy is to train a text encoder to obtain latent codes that align with the hierarchically semantic of the aforementioned pretrained GAN model.
The second strategy is to directly optimize the latent codes in the latent space of the pretrained GAN model with guidance from a pretrained language model.
The latent codes can be randomly sampled from a prior distribution or inverted from a given image, which provides inherent supports for both image generation and manipulation from multi-modal inputs, such as sketches or semantic labels, with textual guidance.
To facilitate text-guided multi-modal synthesis, we propose the \textsc{Multi-Modal CelebA-HQ}, a large-scale dataset consisting of real face images and corresponding semantic segmentation map, sketch, and textual descriptions.
Extensive experiments on the introduced dataset demonstrate the superior performance of our proposed method. 
\thanks{Code and data are available at \protect\url{https://github.com/weihaox/TediGAN}.}

\end{abstract}

\begin{IEEEkeywords}
Generative adversarial network, GAN inversion, text-to-image face image generation, face editing.
\end{IEEEkeywords}

}

%% file: sections/introduction.tex
\IEEEraisesectionheading{\section{Introduction}\label{sec:intro}}

\IEEEPARstart{C}{reating} the desired visual contents without tedious manual operations is a difficult but meaningful task.
To make the process more readily and user-friendly, recent studies have been leveraging a variety of intermediary modalities as conditional guidance, \eg, sketch~\cite{ghosh2019isketchnfill}, semantic label~\cite{isola2017image,wang2018high}, or textual description~\cite{nam2018tagan,xu2018attngan}.
Compared with the success of their label and sketch counterparts, most state-of-the-art methods on text-guided image creation are only able to produce relatively low-quality images~\cite{reed2016generative,dong2017semantic}. 
Those aiming at creating high-quality images with textual guidance typically design a multi-stage architecture and train their models in a progressive manner. 
An initial image with rough shape and color would be refined to a high-resolution one.
To be more specific, there are usually three stages in the main module, and each stage contains a generator and a discriminator. Three stages are trained at the same time, and progressively generate images of three different scales, \ie, \(\text{64}^2 \to \text{128}^2 \to \text{256}^2\).
However, when it comes to higher resolution, the multi-stage training process is time-consuming and cumbersome, making the aforementioned methods unfeasible for high-quality images with \mbox{$\text{1024}^2$} resolution.
More importantly, high-resolution images contain lots of facial details, such as stubble, freckles, or skin pores, and cannot be obtained by simply upsampling from the lower-resolutions or such a progressive training procedure. 
Another serious problem with the previous methods is the poor generalization. Methods trained on small datasets typically fail to handle out-of-distribution data, let alone open-world images and texts
with all sorts of complexities.

\figteaser

\figoverview

\fighighres

Due to the latest developments in generative adversarial networks (GANs), an entirely different image generation paradigm that achieves phenomenal quality, fidelity, and realism has been established.
StyleGAN~\cite{karras2019style,karras2020analyzing}, one of the most notable GAN frameworks, introduces a novel style-based generator architecture and produces images with unmatched resolution and photorealism.
Some recent work~\cite{karras2019style,zhu2020indomain} has demonstrated that the intermediate latent space $\W$ of StyleGAN, inducted from a learned piece-wise continuous mapping, yields less entangled representations and offers more feasible manipulation.
The superior characteristics of $\W$ space and its variants intrigue numerous researchers to develop advanced GAN inversion techniques~\cite{xia2021survey,bau2019inverting} to find the inverted codes for real images in the StyleGAN's latent space and perform meaningful manipulation afterward.
The most popular way~\cite{zhu2020indomain,richardson2020encoding} is to train an additional encoder to map real images into the $\W$ space, which leads to not only faithful reconstruction but also semantically meaningful editing. 
Furthermore, it is easy to introduce the hierarchically semantic property of the $\W$ space to any GAN model by simply learning an extra mapping network.
Such properties inspire us to design cross-modal methods for visual content creation using these fixed, pretrained StyleGAN generators.

In this paper, we propose a very simple yet effective method for \textit{Te}xt-guided \textit{di}verse image generation and manipulation via \textit{GAN} (abbreviated \textit{TediGAN}). 
Our proposed TediGAN, for the first time, unifies the two different tasks, text-guided image generation and manipulation, into one single framework, leading to continuous operations from generation to manipulation and inherent supports for image synthesis from multi-modal inputs, such as sketches or semantic labels with textual guidance, as demonstrated in Figure~\ref{fig:teaser}.
To be specific, we introduces two strategies in this work.
The first strategy, as shown in Figure~\ref{fig:overview} (a) and (b), is to map multi-modal information, including texts, sketches, and labels, into a common latent space of a pretrained StyleGAN. 
Speaking more concretely, we train a encoder for a specific modality that can obtain latent codes that align with the hierarchically semantic of a pretrained GAN model.
It is achieved through three modules.
The first inversion module is to train an image encoder where the inverted code of a given image can be found in the $\W$ space.
The second visual-linguistic similarity module learns linguistic representations that align with the visual representations by projecting both image and text into a common $\W$ space.
The third instance-level optimization module is for identity preservation during editing. It aims to precisely manipulate the desired attributes consistent with the texts while faithfully reconstructing the unconcerned ones.
The second strategy, as shown in Figure~\ref{fig:overview} (c) and (d), optimizes the latent codes directly in the latent space of the aforementioned pre-trained GAN model with the guidance from a pretrained language model.
Like StyleGAN, such off-the-shelf language models~\cite{radford2021learning,jia2021scaling} have helped some cross-modal methods improve performance but are less-addressed (if not never) in text-guided visual content creation. 
Both proposed strategies generate diverse and high-quality results with a resolution up to \mbox{$\text{1024}^2$} and support image creation from multi-modal inputs, such as sketches or semantic labels with textual guidance, as shown in Figure~\ref{fig:high-res-multi-modality}.
Our method inherits the diversity and generalizability of the pretrained StyleGAN model 
and thus provides a reliable guarantee of plausible results no matter how uncommon the given text or image is. 
Compared with the first strategy, the second one can create images from open-world images or texts and manipulate region-of-interest regions for a given image.
Furthermore, to fill the gaps in the text-to-image synthesis dataset for faces, we introduce the \textsc{Multi-Modal CelebA-HQ} dataset to facilitate the research community.
Following the format of the two popular text-to-image synthesis datasets, \ie, CUB~\cite{wah2011caltech} for birds and COCO~\cite{lin2014microsoft} for natural scenes, we create ten unique descriptions for each image in the CelebA-HQ~\cite{karras2017progressive}. 
Besides real faces and textual descriptions, the introduced dataset also contains label maps and sketches for the text-guided generation with multi-modal inputs.

In sum, this paper makes the following contributions.
\begin{itemize} 
\item We propose a unified framework that can generate diverse images given the same input text, or manipulate the given image with a text, allowing the user to edit the appearance of different attributes interactively.
\item We propose two strategies that uses the superior characteristics of a pretrained GAN model for text-guided image generation and manipulation.
Both proposed strategies can create high-quality results, for the first time, with a resolution up to \mbox{$\text{1024}^2$}.
\item We introduce the Multi-Modal CelebA-HQ dataset, which consists of multi-modal face images and corresponding textual descriptions, to facilitate the research community.
\end{itemize}

A preliminary version of this work has been published as a conference paper~\cite{xia2021tedigan}. 
The journal extension improves over the conference paper mainly in two significant ways:
\begin{itemize} 
\item We propose a new strategy 
that uses powerful off-the-shelf language models. Different from and complementary to the initial strategy, it supports image creation from open-world images or texts and region-of-interest manipulation.
\item We investigate in much more detail than the conference version add considerable discusses, such as open-world texts and images. Such extension allow us to further investigate the current limitations and future directions of our proposed method.
\end{itemize}

%% file: sections/related-work.tex
\section{Related Work}
\label{sec:related_work}

\figcompgen

\figcompman

\noindent\textbf{Controllable Image Generation and Manipulation.}
For interactive and controllable image creation, some methods focus on image synthesis conditioned on a variety of user-defined guidance, \eg, label~\cite{zhu2020semantically,zhu2020sean}, sketch~\cite{zou2019language,jo2019sc}, text~\cite{xu2017text,nam2018text}, or control vector like gaze~\cite{xia2020gaze} or light direction~\cite{zhou2019deep}. 
For example, Zhu \etal~\cite{zhu2020semantically} propose a semantically multi-modal image synthesis method that is able to generate diverse images of different attributes from segmentation masks. 
Zhu~\etal~\cite{zhu2020sean} introduce semantic region-adaptive normalization to control the style of each semantic region individually.
Zou~\etal~\cite{zou2019language} propose a network that colourizes sketches following the instructions provided by the input text specifications. 
Jo~\etal~\cite{jo2019sc} propose a face editing method where users can edit face images using sketch and color.
Both sketch-based and label-based categories put forward high requirements for the user's drawing~\cite{xia2021sketch}.
It is challenging to synthesize natural and realistic images from poorly-drawn sketches or labels. 
Those conditioned on control vectors are typically limited to a small range of applications such as gaze redirection~\cite{xia2020gaze} or relighting~\cite{sun2019single,zhou2019deep}.
Although some methods, including both sketch-based~\cite{jo2019sc} and label-based~\cite{wang2018high,zhu2020sean} ones, provide an interactive software for users to edit images, some important attributes such as color and texture can only get random results. Compared to providing a color or texture gallery, a more feasible way is to integrate descriptive texts.
In general, compared with labels, sketches, and vectors, texts have a much lower cost of learning for users to edit real images. 
Based on the pros and cons of these user-defined conditions, sketch or label is integrated in our application to define the basic features of a human face like head pose or face shape, \etc, while text precisely defines color, texture, and other desired attributes.

\noindent\textbf{Text-to-Image Generation.}
There are primarily two categories of GAN-based text-to-image generation methods.
The first category produces images from texts directly by one generator and one discriminator.
For example, Reed~\etal~\cite{reed2016generative} propose to use conditional GANs to generate plausible images from given text descriptions. 
Tao~\etal~\cite{tao2020dfgan} propose a simplified backbone that generates high-resolution images directly by Wasserstein distance and fuses the text information into visual feature maps to improve the image quality and text-image consistency.
Despite the plainness and conciseness, the one-stage models produce dissatisfied results in terms of both photo-realism and text-relevance in some cases.
Thus, another thread of research focuses on multi-stage processing.
Zhang~\etal~\cite{zhang2017stackgan} stack two GANs to generate high-resolution images from text descriptions through a sketch-refinement process.
They further propose a three-stage architecture~\cite{zhang2018stackgan++} that stacks multiple generators and discriminators, where multi-scale images are generated progressively in a course-to-fine manner.
Xu~\etal~\cite{xu2018attngan} improve the work of~\cite{zhang2018stackgan++} from two aspects. 
First, they introduce attention mechanisms to explore fine-grained text and image representations.
Second, they propose a Deep Attentional Multimodal Similarity Model (DAMSM) to compute the similarity between the generated image and the sentence.
The subsequent studies basically follow the framework of~\cite{xu2018attngan} and have proposed several variants by introducing different mechanisms like attention~\cite{li2019control} or memory writing gate~\cite{zhu2019dmgan}.
However, the multi-stage frameworks produce results that look like a simple combination of visual attributes from different image scales.
A concurrent zero-shot text-to-image system called DALL-E~\cite{ramesh2021zero} is trained on large-scale unconstrained image-text pairs and can generate arbitrary classes of images from open-world descriptions.
Despite the ability to generate arbitrary classes of images from open-world descriptions, the quality and resolution of its results are often limited compared with the state-of-the-art text-to-image methods that focus on certain classes~\cite{bau2021paint}.

\noindent\textbf{Text-Guided Image Manipulation.}
Text-guided image manipulation is similar to text-to-image generation in terms of producing results that contain desired visual attributes. 
Differently, the modified results should only change certain parts and preserve text-irrelevant contents of the original images. 
For example, Dong~\etal~\cite{dong2017semantic} propose an encoder-decoder architecture to modify an image according to a given text.
Nam~\etal~\cite{nam2018text} disentangle different visual attributes by introducing a text-adaptive discriminator, which can provide finer training feedback to the generator. 
Li \etal~\cite{li2020manigan} introduce a multi-stage network with a novel text-image combination module to produce high-quality results.
Liu \etal~\cite{liu2020describe} propose a multi-domain and multi-modal method to explicitly model the visual attributes of an image and learn how to translate them through automatically generated commands.
Similar to text-to-image generation, the text-based image manipulation methods with the best performance are basically based on the multi-stage framework.
Different from all existing methods, we propose a novel framework that unifies text-guided image generation and manipulation methods and can generate high-resolution and diverse images \texttt{directly} without multi-stage processing.

\noindent\textbf{Vision-language Representation Learning.}
One key of text-guided image synthesis is to match visual attributes with corresponding words. To do this, current methods usually provide explicit word-level training feedback from the elaborately-designed discriminator~\cite{li2020manigan,li2020lightweight}.
There is also a rich line of work proposed to address a related direction named image-text matching, or visual-semantic alignment, aiming at exploiting the matching relationships and making the corresponding alignments between text and image.
Most of them can be categorized into two-branch deep architecture according to the granularity of representations for both modalities, \ie, global~\cite{mao2014deep,ma2015multimodal} or local~\cite{karpathy2015deep} representations.
The first category employs deep neural networks to extract the global features of both modalities, based on which their similarities are measured~\cite{mao2014deep}.
Another thread of work performs instance-level image-text matching~\cite{lee2018stacked,song2019polysemous}, learning the correspondences between words and image regions.
Pre-training has also been an increasingly significant approach in vision-language modeling, especially those on millions or billions of text-image pairs collected from a variety of publicly available sources on the Internet~\cite{jia2021scaling,radford2021learning}. 
For example, 
Radford~\etal~\cite{radford2021learning} propose an efficient method of learning from natural language supervision, called Contrastive Language-Image Pre-training (CLIP), which is trained on 400 million text-image pairs. 
By contrast, the dataset in~\cite{jia2021scaling} is much larger (1.8B image-text pairs) and noisier. They propose a simple dual-encoder architecture using a contrastive loss that avoids heavy labor on data curation and requires only minimal frequency-based cleaning.
Both methods have shown strong potential on the semantic similarity estimation between given texts and images.

%% file: sections/method-A.tex
\section{The TediGAN Framework}
\label{sec:method}

In this section, we introduce the two proposed strategies for text-guided diverse face image generation and manipulation. 
In Section~\ref{subsec:gan-inversion}, we introduce the inversion module used in both strategies, \ie, training an image encoder to map the real images to the latent space such that all codes produced by the encoder can be recovered at both the pixel-level and the semantic-level. 
In Section~\ref{subsec:train-text-encoder}, we demonstrate how to use the hierarchical characteristic of $\W$ space to train a text encoder. 
We propose the visual-linguistic similarity learning (Section~\ref{subsec:vls-model}) that maps the image and text into the same joint embedding space.
To preserve identity in manipulation, we propose an instance-level optimization (Section~\ref{subsec:instance-level-optimization}), involving the trained encoder as a regularization to better reconstruct the pixel values without affecting the semantic property of the inverted code.
In Section~\ref{subsec:pretrained-text-encoder}, we demonstrate how to achieve the same goal through a pretrained powerful language model and provide some necessary discusses and innovative extensions. We extend our method to enable users to generate or manipulate via arbitrary descriptions of facial attributes and concentrate on manipulation in the region of interest.

\subsection{StyleGAN Inversion Module}
\label{subsec:gan-inversion}
The inversion module aims at training an image encoder that can map a real face image to the latent space of a fixed StyleGAN model pretrained on the FFHQ dataset~\cite{karras2019style}.
The reason we invert a trained GAN model instead of training one from scratch is that, in this way, we can go beyond the limitations of a paired text-image dataset. 
The StyleGAN is trained in an unsupervised setting and covers much higher quality and wider diversity, leading to a latent space that almost covers the full space of facial attributes, which makes our method able to produce satisfactory edited results with images in the wild. 
In order to facilitate subsequent alignment with text attributes, our goal for inversion is not only to reconstruct the input image by pixel values but also to acquire the inverted code that is semantically meaningful and interpretable~\cite{shen2020interpreting}.

Before introducing our method, we first briefly establish problem settings and notations.
A GAN model typically consists of a generator $G(\cdot): \Z\rightarrow\X$ to synthesize fake images and a discriminator $D(\cdot)$ to distinguish real data from the synthesized. 
In contrast, GAN inversion studies the reverse mapping, which is to find the best latent code $\z^{*}$ by inverting a given image $\x$ to the latent space of a well-trained GAN.
A popular solution is to train an additional encoder $E_v(\cdot): \X\rightarrow\Z$ (subscript $v$ means visual).
To be specific, a collection of latent codes $\z^{s}$ are first randomly sampled from a prior distribution, \eg, normal distribution, and fed into $G(\cdot)$ to get the synthesis $\x^{s}$ as the training pairs.
The introduced encoder $E_v(\cdot)$ takes $\x^{s}$ and $\z^{s}$ as inputs and supervisions respectively and is trained with
\begin{align}
\min_{\Theta_{E_v}}\Loss_{E_v} = ||\z^{s} - E_v(G(\z^{s}))||_2^2, 
\label{eq:conventional-encoder}
\end{align}
where $||\cdot||_2$ denotes the $l_2$ distance and $\Theta_{E_v}$ represents the parameters of the encoder $E_v(\cdot)$.

Despite of its fast inference, the aforementioned procedure simply learns a deterministic model with no regard to whether the codes produced by the encoder align with the semantic knowledge learned by $G(\cdot)$.
The supervision by only reconstructing $\z^{s}$ is not powerful enough to train $E_v(\cdot)$, and $G(\cdot)$ is actually not fully used to guide the training of $E_v(\cdot)$, leading to the incapability of inverting real images.
To solve these problems, we use a totally different strategy to train an encoder for GAN inversion as in~\cite{zhu2020indomain}. 
There are two main differences compared with the conventional framework:
(a) the encoder is trained with real images rather than with synthesized images, making it more applicable to real applications;  
(b) the reconstruction is at the image space instead of latent space, which provides semantic knowledge and accurate supervision and allows integration of powerful image generation losses such as perceptual loss~\cite{johnson2016perceptual} and LPIPS~\cite{zhang2018unreasonable}.
Hence, the training process can be formulated as
\begin{small}
\begin{align}
  &\begin{aligned}
    \min_{\Theta_{E_v}}\Loss_{E_v} \!=\! ||\x - G(E_v(\x))||_2^2\ &\!+\!\lambda_{1} ||F(\x) \!-\! F(G(E_v(\x)))||_2^2\ \\
    &\!-\!\lambda_{2}\E[D_v(G(E_v(\x)))], \label{eq:encoder}
  \end{aligned} \\
  &\begin{aligned}
    \min_{\Theta_{D_v}}\Loss_{D_v} = \!\E[D_v(G(E_v(\x)))] \!- \!\E[D_v(\x)] \! + \!\frac{\lambda_3}{2} {\E}[||\nabla_{{\x}}D_v(\x)||_2^2],
  \end{aligned} \label{eq:discriminator}
\end{align}
\end{small}
where $\Theta_{E_v}$ and $\Theta_{D_v}$ are learnable parameters, $\lambda_{1}$ and $\lambda_{2}$ are the perceptual and discriminator loss weights, $\lambda_{3}$ is the hyper-parameter for the gradient regularization, and $F(\cdot)$ denotes the VGG feature extraction model.

Through the learned image encoder, we can map a real image into the $\mathcal{W}$ space and obtain a latent code. 
The obtained code is guaranteed to align with the semantic domain of the StyleGAN generator and can be further utilized to mine cross-modal similarity between the image-text instance pairs.

\subsection{Training an Text Encoder}
\label{subsec:train-text-encoder}

\subsubsection{Visual-Linguistic Similarity Learning}
\label{subsec:vls-model}

Once the inversion module is trained, given a real image, we can map it into the $\mathcal{W}$ space of StyleGAN. 
The next problem is how to train a text encoder that learns the associations and alignments between image and text.
The previous methods usually use a submodule named Deep Attentional Multimodal Similarity Model (DAMSM)~\cite{xu2018attngan,li2019control,li2020manigan,liang2020cpgan,tao2020dfgan} as the text-image matching loss, which learns two neural networks that map subregions of the image and words of the sentence to a common semantic space, thus measures the image-text similarity at the word level to compute a fine-grained loss for image generation.
However, this module is cumbersome and fails to explore the relationships between face attributes and sentence fragments, leading to entangled attributes and limited matching accuracy.

Instead of training a text encoder in the same way as the image encoder or the aforementioned DAMSM, we propose a visual-linguistic similarity module to project the image and text into a common embedding space, \ie, the $\W$ space, as shown in Figure~\ref{fig:overview} (a).
Given a real image and its descriptions, we encode them into the $\W$ space by using the previously trained image encoder and a text encoder. 
The obtained latent code is the concatenation of $L$ different $C$-dimensional $\w$ vectors, one for each input layer of StyleGAN.
The multi-modal alignment can be trained with 
\begin{align}
  \min_{\Theta_{E_l}}\Loss_{E_l} = ||\sum_{i=1}^{L} p_i (\w^v_i - \w^l_i)||_2^2, 
  \label{eq:reg}
\end{align}
where $\Theta_{E_l}$ represents the parameters of the text encoder $E_l(\cdot)$ and subscript $l$ means linguistic;
$\w^v, \w^l \in \W^{L \times C}$ are the obtained image embedding and text embedding; 
$\w^v = f(E_v(\x))$ is the projected code of the image embedding $\z$ in the input latent space $\Z$ using a non-linear mapping network $f:\Z \to \W$; $\w^l$ shares a similar definition;
$\w^v$ and $\w^l$ are with the same shape $L \times C$, meaning to have $L$ \textit{layers} and each with a $C$-\textit{dimensional} latent code;
and $p_i$ is the weight of $i$-th layer in the latent code.

Compared with DAMSM, our proposed module is lightweight and easy to train. More importantly, this module achieves instance-level alignment, \ie, learning correspondences between visual and linguistic attributes, by leveraging the disentanglability of StyleGAN.
The text encoder is trained with the proposed visual-linguistic similarity loss together with the pairwise ranking loss~\cite{dong2017semantic}, which is omitted from Equation~\ref{eq:reg}.

\figcontrol

\figlayer

\subsubsection{Instance-Level Optimization}
\label{subsec:instance-level-optimization}
One of the main challenges of face manipulation is the identity preservation. Due to the limited representation capability, learning a perfect reverse mapping with an encoder alone is not easy. 
To preserve identity, some recent methods~\cite{richardson2020encoding,chai2021using} incorporate a dedicated face recognition loss~\cite{deng2019arcface} to measure the cosine similarity between the output image and its source.
Different from their methods, for text-guided image manipulation, we implement an instance-level optimization module to precisely manipulate the desired attributes consistent with the descriptions while faithfully reconstructing the unconcerned ones.  
We use the inverted latent code $\z$ as the initialization, and the image encoder is included as a regularization to preserve the latent code within the semantic domain of the generator. 
To summarize, the objective function for optimization is
\begin{align}
  \begin{aligned}
    \z^{*} = \arg\min_{\z}\ ||\x - G(\z)||_2^2\ &+ \lambda_{1}^{\prime}||F(\x) - F(G(\z))||_2^2\ \\
    &+ \lambda_{2}^{\prime}||\z - E_v(G(\z))||_2^2,
  \end{aligned} \label{eq:optimization}
\end{align}
where $\x$ is the original image to manipulate. $\lambda_{1}^{\prime}$ and $\lambda_{2}^{\prime}$ are hyperparameters corresponding to the perceptual loss and the encoder regularization term, respectively.

\subsubsection{Control Mechanism}
\label{subsec:control_mechanism}
In this section, we demonstrate how the two different tasks (text-to-image generation and text-guide image manipulation) and different modalities (image, text, sketch, and label) can be unified into one framework by a simple control mechanism. 

\noindent\textbf{Attribute-Specific Selection.} 
Our control mechanism is based on the style mixing of StyleGAN.
The layer-wise representation of StyleGAN learns disentanglement of semantic fragments (attributes or objects).
In general, different layer $\w_i$ represents different attributes and is fed into the $i$-th layer of the generator. 
Changing the value of a certain layer would change the corresponding attributes of the image.
As shown in Figure~\ref{fig:control_mechanism}, given two codes with the same size $\w^c, \w^s \in \W^{\;L \times C}$ denoting content code and style code, this control mechanism selects attribute-specific layers and mixes those layers of $\w^s$ by partially replacing corresponding layers of $\w^c$. 
For text-to-image generation, the produced images should be consistent with the textual description, thus $\w^c$ should be the linguistic code, and randomly sampled latent code with the same size acts as $\w^s$ to provide diversity.
For text-guided image manipulation, $\w^c$ is the visual embedding while $\w^s$ is the linguistic embedding;
the layers for mixing should be relevant to the text, for the purpose of modifying the relevant attributes only and keeping the unrelated ones unchanged.

\noindent\textbf{Supported Modality.} 
The style code $\w^s$ and content code $\w^c$ could be sketch, label, image, and noise, as shown in Figure~\ref{fig:control_mechanism}, which makes our TediGAN feasible for multi-modal image synthesis.
The control mechanism provides high accessibility, diversity, controllability, and accurateness for image generation and manipulation.
Due to the control mechanism, as shown in Figure~\ref{fig:teaser}, our method inherently supports continuous operations and multi-modal synthesis for sketches and semantic labels with descriptions.
To produce the diverse results, all we need to do is to keep the layers related to the text unchanged and replace the others with the randomly sampled latent code.
If we want to generate images from other modality with text guidance, take the sketch as an example, we can train an additional sketch image encoder $E_{vs}$ in the same way as training the real image encoder and leave the other parts unchanged.

\noindent\textbf{Layerwise Analysis.} 
For Style-based generators, the image size $\mathrm{R}$ is determined by its number of layers $L$: $L\!=\!2\log_2\mathrm{R}\!-\!2$.
For example, the pre-trained StyleGAN we used in most experiments is to generate images of size 256 $\times$ 256, which has 14 layers of the intermediate vector.
For a synthesis network trained to generate images of 512 and 1024, the intermediate vector would be of shape (16, 512) and (18, 512).
In general, layers in the generator at lower resolutions control high-level styles such as eyeglasses and head pose,
layers in the middle control hairstyle and facial expression, while the final layers control color schemes and fine-grained details.
In order to find the correspondence between layers and attributes, we conduct a layerwise analysis by selecting some images with different attributes and performing layer-wise style mixing.
The results are shown in Figure~\ref{fig:layerwise_analysis}.
The top figure shows the results of replacing the top-$n$ layers while the bottom demonstrates the results of replacing one specific layer.
The real images in the green rectangle are illustrated for comparison with the ones after style mixing.
Four sets of images were generated from their respective modality (source content and source style); the rest images were generated by copying a specified subset of styles from style images and taking the rest from the content ones.
It is obvious that each layer in StyleGAN roughly controls certain attributes. 
For example, as illustrated, replacing the first layer of the inverted latent code of Emma Watson with the counterpart of Yann LeCun adds a pair of eyeglasses for her. 
The layers from 11-14 represent micro features or fine structures, such as stubble, freckles, or skin pores, which can be regarded as the stochastic variation.
High-resolution images contain lots of facial details and cannot be obtained by simply upsampling from the lower-resolutions, making the stochastic variations especially important as they improve the visual perception without affecting the main structures and attributes of the synthesized image.
Based on empirical observations, we list some attributes represented by different possible layers of a 14-layer StyleGAN in Table~\ref{tab:layerwise_analysis}. 
When performing style mixing for target attributes, one can refer to Table~\ref{tab:layerwise_analysis} or other layerwise analysis~\cite{shen2020interpreting} as a start.

\figdata

\tablayer

\figdiverse

%% file: sections/method-B.tex
\subsection{Using A Pretrained Text Encoder}
\label{subsec:pretrained-text-encoder}

\subsubsection{An optimization problem}
In Section~\ref{subsec:vls-model}, we introduce a strategy that trains a text encoder to produce text codes sharing the same structure of image codes produced by an image encoder.
The improvement of GAN generator goes hand in hand with progress of language models. 
Since we use a pre-trained generator, what if using a pre-trained language model instead of training a text encoder from scratch? 
In this section, we demonstrate a simple strategy that feasibly introduces some powerful pretrained language models, \eg, CLIP~\cite{radford2021learning} or ALIGN~\cite{jia2021scaling}, to replace the visual-linguistic similarity learning module.
We take CLIP~\cite{radford2021learning} as an example. 
The CLIP model jointly trains an image encoder and a text encoder to predict the correct pairings of a batch of image and text training samples.
At test time, given a image and a text, the CLIP encoder computes the cosine similarity $S(\x,t)$ between encoded features of given image $\x$ and text $t$.

In this case, we have the StyleGAN model $G$ pretrained on FFHQ and the pretrained text encoder CLIP \texttt{ViT/32B} text-image semantic similarity network $C$.
The goal is two-fold: we want the generated images or manipulated attributes of given images to be firstly visually satisfactory and secondly semantically consistent with the given texts.
For image synthesis using a pretrained generator and a pretrained text encoder for the two-fold goal, the most intuitive way is to solve the following optimization problem:
\begin{align}
  \begin{aligned}
    \z^{*} = \arg\min_{\z}\ ||\z - E_v(\x)||_2^2+ \lambda \Loss_{\text{CLIP}}(\x,t),
  \end{aligned} 
  \label{eq:optimization_2}
\end{align}
where $\x=G(\z)$ and  $\Loss_{\text{CLIP}}(\x,t)=1-S(\x,t)$. 
For manipulation, given an image $\x$, we obtain its inverted latent code $\z$ as the initialization for the optimization, using the inversion module in Section~\ref{subsec:gan-inversion} or other available inversion methods~\cite{richardson2020encoding,chai2021using}. 
For image generation, we use the randomly sampled or mean codes as the initialization.

\subsubsection{Parameter Sensitivity}
Generally, the last term in Equation~\eqref{eq:optimization_2} is for the semantic consistency between the generated image and the given text, while the other forces the unrelated attributes or regions unchanged and produces results with a plausible appearance. 
However, Equation~\eqref{eq:optimization_2} is very sensitive to the parameter. Different edits require different parameters of the semantic consistent term and it is hard to decide what $\lambda$ to use before we start. 
One possible reason for this phenomenon, which we call parameter sensitivity, is that both pretrained models $G$ and $C$ are trained for different tasks and it is hard to find the exact optimum for both models when optimizing a single instance.
One solution is to add additional image reconstruction term and perceptual term to help stabilize the optimization process:
\begin{align}
  \begin{aligned}
    \z^{*} =&\arg\min_{\z}\ ||\x - G(\z)||_2^2\ + \lambda_{1}^{\prime}||F(\x) - F(G(\z))||_2^2\ \\
    &+ \lambda_{2}^{\prime}||\z - E_v(G(\z))||_2^2+ \lambda_{3}^{\prime}\Loss_{\text{CLIP}}(\x,t).
  \end{aligned} 
  \label{eq:optimization_3}
\end{align}
This optimization can be seen a revised version of instance-level optimization introduced in Section~\ref{subsec:instance-level-optimization} by adding a text-image semantic consistent term.
The original latent term and the two additional reconstruction and perceptual terms in Equation~\eqref{eq:optimization_3} make the image reconstruction higher priority than the goal of making textually consistent changes.
We found in experiments that the aforementioned improvements decrease the parameter sensitivity and most cases can be well-handled using one fixed parameter setting.
Another solution lies in the choices of optimizers. 
We use the gradient-based algorithm Adam~\cite{kingma2014adam}.
Besides gradient descent, some other optimization strategies like Covariance Matrix Adaptation (CMA)~\cite{hansen2001cma} can be used to find optimal results for both models $G$ and $C$.
Some concurrent GAN inversion methods~\cite{huh2020transforming} demonstrate that gradient-free optimizers find better solutions than gradient-based methods.

\subsubsection{Region-of-interest Manipulation}

Such optimization can be further extended to the region-of-interest manipulation, which focuses the optimization on the selected region~\cite{abdal2020image2stylegan2,chai2021using,bau2021paint}. 
Similar to the inversion module introduced in Section~\ref{subsec:gan-inversion}, we train another image encoder, which takes a masked image $\x \otimes m$ and a mask $m$, instead of taking an image $\x$ as input:
\begin{align}
\begin{aligned}
\min_{\Theta_{E_m}}\Loss_{E_{m}} &= ||\x - \x_r||_2^2
+ \lambda_1^{\prime}||F(\x)-F(\x_r)||\\ &+\lambda_2^{\prime}||\z-E_m(\x_m)||,
\end{aligned} 
\label{eq:encoder_mask}
\end{align}
where $\x=G(\z)$, $\x_r=G(E_m(\x_m))$, $\x_m = (\x \otimes m, m)$, $m$ is the mask, and $(\cdot,\cdot)$ operator means channel concatenation. 

The main difference between Equation~\eqref{eq:encoder} and  Equation~\eqref{eq:encoder_mask} is obvious: instead of taking only the masked image as input, $E_{m}$ takes both the masked image $\x \otimes m$ and the mask $m$.
The reason is that, with this mask, we can tell the encoder explicitly which regions are ``unknown'' and unsupervisedly learn an inherent image representation that allows completion from only partial observations.
Given only partial images as input, the encoder is encouraged to simply fit the values of these unknown pixels from the known pixels and produce unpleasing results.
The encoder $E_{m}$ provides additional flexibility, allowing the generator to create user-specified visual contents in the region of interest while leaving the unrelated parts of the image unchanged.
Once the image encoder is trained, given a source image $\x$, a mask $m$ and a description $t$, we can manipulate the masked region with given text using
\begin{align}
\begin{aligned}
\z^{*} &\!=\! \arg\min_{\z}\ ||\x \!\otimes\! m \!-\! \x_r \!\otimes\! m)||_2^2
\!+\! \lambda_1^{\prime}||F(\x)\!-\!F(\x_r)||\\ &+\lambda_2^{\prime}||\z-E_m(\x_m)|| + \lambda_{3}^{\prime}\Loss_{\text{CLIP}}(\x_r,t).
\end{aligned} 
\label{eq:optimization_4}
\end{align}

%% file: sections/experiment.tex
\section{Experiments}
\label{sec:experiments}

\figoodsketch
\tabquangen
\tabquanman

\subsection{Experiments Setup}
\label{subsec:setup}

\noindent\textbf{Multi-Modal CelebA-HQ Dataset.}
To facilitate text-guided multi-modal synthesis, we propose the \textsc{Multi-Modal-CelebA-HQ}, a large-scale face image dataset that has 30,000 high-resolution face images by following CelebA-HQ~\cite{karras2017progressive}. 
Each image has a high-quality semantic segmentation map, a sketch, a descriptive text, and an image with transparent background, as shown in Figure~\ref{fig:sample_data}.
For text descriptions, following the format of the popular CUB~\cite{wah2011caltech} and COCO~\cite{lin2014microsoft} datasets, we create ten unique single sentence descriptions for each image in CelebA-HQ~\cite{karras2017progressive}. The CelebA-HQ dataset consists of facial images and their attributes, where the attribute labels are used to generate natural sounding textual descriptions. 
The total dataset consists of 30000 images, which we divide into 80\% training and 20\% test samples.
A similar dataset~\cite{stap2020conditional} is also proposed but it is not publicly available. 
For labels, we use the labels from CelebAMask-HQ~\cite{CelebAMask-HQ}, which contains manually-annotated semantic mask of facial attributes corresponding to CelebA-HQ. 
To produce a sketch for each image, we first apply Photocopy filter in Photoshop to extract edges, which preserves facial details and introduces excessive noise. We then apply the sketch-simplification~\cite{simoserra2016simplify} to get edge maps resembling hand-drawn sketches. The data generation pipeline is the same as in~\cite{chen2020DeepFaceDrawing,richardson2020encoding}. 
We also provided each image with transparent background. 
For background removing, we use an open-source tool Rembg~\cite{rembg} and a commercial software remove.bg~\cite{remove.bg}. Different backgrounds can be further added using image composition or harmonization methods like DoveNet~\cite{cong2020doveNet}.
The introduced \textsc{Multi-Modal-CelebA-HQ dataset} can be used to train and evaluate algorithms of text-to-image-generation, text-guided image manipulation, sketch-to-image generation, and other various GANs for face generation and editing.

\noindent\textbf{Baseline Models and Evaluation Metric.} 
We evaluate our proposed method on text and image partitions, comparing with state-of-the-art approaches AttnGAN~\cite{xu2018attngan}, ControlGAN~\cite{li2019control}, DM-GAN~\cite{zhu2019dmgan}, and DFGAN~\cite{tao2020dfgan} for text-to-image generation, and comparing with ManiGAN~\cite{li2020manigan} for text-guided image manipulation. 
All methods are retrained with the default settings on the proposed Multi-Modal CelebA-HQ dataset.
We did not compare with some concurrent methods~\cite{patashnik2021styleclip,bau2021paint,ramesh2021zero} since either they have compared with our method~\cite{patashnik2021styleclip} or their implementation is not publicly available~\cite{bau2021paint,ramesh2021zero}.
For evaluation, there are four equally important aspects: quality, diversity, accuracy, and realism.
Following the previous methods~\cite{li2019control,li2020lightweight}, we evaluate the quality of generated or manipulated images using Fr\'echet Inception Distance (FID)~\cite{heusel2017gans}.
The image diversity is measured by the Learned Perceptual Image Patch Similarity (LPIPS)~\cite{zhang2018unreasonable}. 
The accuracy and realism are evaluated through a user study.
For realism, users are asked to judge which one is more photo-realistic among the given results of the aforementioned methods.
The accuracy of generation is evaluated by the similarity between the given description and the corresponding generated image, where users are asked to judge which image is more coherent with the given text.
To evaluate the accuracy of manipulation, besides the modified visual attributes of the synthetic image are aligned with the text, users are also asked to judge whether the text-irrelevant contents are preserved.
We test accuracy and realism by randomly sampling 50 images with the same conditions and collect more than 20 surveys from different people with various backgrounds.

\figclip

\subsection{Network Architecture}
\label{subsec:network}

For experiment in Section~\ref{subsec:train-text-encoder}, we use the a 14-layer StyleGAN generator as the pretrained GAN model for inversion. The image encoders are based on~\cite{zhu2020indomain}.  
The text encoder is based on~\cite{xu2018attngan} with an additional non-linear mapping component the same as in the image encoders.
We use the recently proposed BERT model~\cite{devlin2018bert} instead of the original LSTM~\cite{hochreiter1997lstm} in~\cite{xu2018attngan} to improve text embedding. 
For experiments in Section~\ref{subsec:pretrained-text-encoder}, we use the CLIP as the pretrained text encoder. 
For high-resolution results in both parts, we use pSp~\cite{richardson2020encoding} and StyleGAN2~\cite{karras2020analyzing} as encoder and generator, respectively.

\subsection{Training Details} 
\label{subsec:training}

We first train the StyleGAN inversion module. Once the image encoder learns how to invert a real image into the StyleGAN latent space~\cite{karras2019style}, we then train the visual-linguistic similarity module to make the text and image latent code match.
For StyleGAN inversion training, We basically follow the framework of Zhu~\etal~\cite{zhu2020indomain}.
The StyleGAN model we used for inversion is trained on FFHQ dataset \cite{karras2019style}, which contains 70,000 high-quality face images, and the latent codes for different layers are different. 
When training the image encoder $E_v$, the generator is \emph{fixed} and we only update the encoder and discriminator according to Equation~\eqref{eq:encoder} and Equation~\eqref{eq:discriminator}.
As for the perceptual loss in Equation~\eqref{eq:encoder}, we take $\mathtt{conv4\_3}$ as the VGG output.
The hyper-parameters $\lambda_{1}$ in Equation~\eqref{eq:encoder} and $\lambda_{1}^{\prime}$ in Equation~\eqref{eq:optimization} and Equation~\eqref{eq:optimization_2} are set as $5e^{-5}$, $\lambda_{2}=0.1$, $\lambda_{2}^{\prime}=2$ and $\lambda_{3}$ in Equation~\eqref{eq:discriminator} is $10$. $\lambda_{3}^{\prime}$ in Equation~\eqref{eq:optimization_3} is set as $200$ in most cases.
For visual-linguistic similarity module, we fix the image encoder and train the text encoder using the images and texts in the introduced dataset.

\subsection{Comparison with State-of-the-Art Methods}
\label{subsec:comparison}

\subsubsection{Text-to-Image Generation}
\label{subsec:exp_gen}

\noindent\textbf{Quantitative Comparison.}
In our experiments, we evaluate the FID and LPIPS on a large number of samples generated from randomly selected text descriptions. 
To evaluate accuracy and realism, we generate images from 50 randomly sampled texts using different methods. 
In a user study for accuracy, users are asked to judge which one is the most coherent with the given texts while in another for realism to choose the most photo-realistic image.
We use \textit{TediGAN-A} and \textit{TediGAN-B} to refer to the two strategies proposed in Section~\ref{subsec:train-text-encoder} and Section~\ref{subsec:pretrained-text-encoder}.
The results are demonstrated in Table~\ref{tab:quan_gen}. 
Compared with the state-of-the-arts, our method achieves better FID, LPIPS, accuracy, and realism values, which proves that our methods can generate images with the highest quality, diversity, photorealism, and text-relevance. 

\noindent\textbf{Qualitative Comparison.} 
Figure~\ref{fig:comp_gen} demonstrates the visual comparisons of different text-to-image generation methods and our two proposed strategies. 
As shown, most methods can generate photo-realistic and text-relevant results. 
When comparing comprehensively, however, the gap of performance is obvious.
For example, some of the attributes contained in the text sometimes do not appear in the generated image, \eg, lipstick for AttnGAN, gender for DF-GAN, and hairstyle for ControlGAN.
The generated images look like featureless paints and lack essential details that make them indistinguishable from the real images.
This ``featureless painterly'' appearance~\cite{karras2019style} would be significantly obvious and irredeemable when generating higher resolution images using the multi-stage training methods~\cite{xu2018attngan,li2019control,zhu2019dmgan} due to their incompetent to generate facial details like stubble, freckles, or skin pores.
Furthermore, most existing solutions have limited diversity of the outputs, even if the provided conditions contain different meanings. For example, ``\textit{has a beard}" might mean a goatee, short or long beard, and could have different colors.
Our method can not only generate results with diversity but also realise the expectation to change where you want by using the control mechanism.
To produce diverse results, with the layers related to the text unchanged, the other layers could be replaced by any values sampled from the prior distribution. 
For example, as shown in the first row of Figure~\ref{fig:diverse_image}, the key visual attributes (\textit{women, black long hair, earrings, and smiling}) are preserved, while the other attributes, like haircuts, makeups, face shapes, and head poses, show a great degree of diversity.
The images in the second row illustrate more precise control ability. We keep the layers representing face shape and head pose the same and change the others. 
Our method is also able to generate high-quality and diverse results with resolution at 1024 $\times$ 1024 from multimodal inputs, \eg, free-drawn sketches or semantic labels with textual descriptions. The results are shown in Figure~\ref{fig:high-res-multi-modality}.

\subsubsection{Text-Guided Image Manipulation}
\label{subsec:exp_man}

\noindent\textbf{Quantitative Comparison.}
In our experiments, we evaluate the FID score and conduct a user study on accuracy and realism by selecting images randomly from both \textit{CelebA} and \textit{Non-CelebA} datasets with randomly chosen descriptions from our proposed dataset and some open-world descriptions (marked as \textit{Open-Text}). 
The quantitative results are shown in Table~\ref{tab:quan_man}. 
Compared with ManiGAN~\cite{li2020manigan}, both of our proposed strategies achieve better FID, accuracy, and realism. 
This indicates that our method can produce high-quality synthetic images, and the modifications are highly aligned with the given descriptions, while preserving other text-irrelevant contents. 

\figgenopen

\noindent\textbf{Qualitative Comparison.} 
Figure~\ref{fig:comp_man} shows the visual comparisons between a recent method ManiGAN~\cite{li2020manigan} and our two proposed strategies, \textit{TediGAN-A} and \textit{TediGAN-B}.
ManiGAN produces less satisfactory modified results: the text-relevant regions are not modified and the text-irrelevant ones are changed. 
In some cases, especially for the difficult attributes and out-of-distribution images, the obtained results are basically the replications of the original images with worse quality.
For example, the second row is the results of adding earrings and changing the face shape and hair style of the pictured woman. Our method completes this difficult case while ManiGAN fails to produce required attributes.
The images in last two columns are results of out-of-distribution (Non-CelebA), \ie, images from other face dataset such as~\cite{chelnokova2014rewards, courset2018caucasian, yi2019apdrawinggan}, which illustrate that our method is prepared to produce pleasing results for the in-the-wild images.
Compared with \textit{TediGAN-A}, which is the initial strategy proposed in the conference version~\cite{xia2021tedigan}, the newly proposed \textit{TediGAN-B} can further handle image creation from open-world images or texts and region-of-interest manipulation, which will be discussed in Section~\ref{subsec:open-world} and Section~\ref{subsec:roi}.

\subsection{Diving Deep into Open-World Images and Texts}
\label{subsec:open-world}

In this section, we illustrate the results of images and texts with real-world variances.
The StyleGAN we used is pretrained on a very large face dataset~\cite{karras2019style}. 
Although the StyleGAN model is trained unsupervisedly, its latent space almost covers the full space of facial attributes, where we can discover the interpretable directions with text guidance. 
With the benefit of the powerful language model pretrained on large-scale image and text pairs, our method~\textit{TediGAN-B} can create image from open-world texts or manipulate open-world face images with open-world descriptions. 
It is surprised to see that although the $W$-space is never trained or fine-tuned with the additional texts, we can still obtain the desired latent code that corresponds with the given open-world descriptions.

The robust encoders support inverting open-world images, labels, and sketches into the latent space of StyleGAN and enables us to synthesis images from different kinds of images or sketches. The results from texts and sketches of different styles are shown in Figure~\ref{fig:ood_sketch}. 
The sketches in the first row are real human-drawn sketches. The second are different degradations of one sample randomly chosen from our sketch data, from top to bottom adding noisy scribbles, mask, colorful text, and irregular shape. Our method still produces plausible results despite that the model has not seen these difficult variants during training.

In Figure~\ref{fig:with-clip}, we show some manipulated results by our TediGAN-B.
The first row of images shows manipulated results using texts in (\textit{makeup}, \textit{young}, and \textit{grey hair}) and out of (\textit{asia} and \textit{tanned}) the proposed dataset while the second are manipulated with similar descriptions of \textit{beardless}.
The results of image generation with open-world texts can be found in Figure~\ref{fig:gen_open}. 
The images in each column are obtained from the same randomly-sampled latent code according to different texts. Each description has the form of "he/she is/has/looks like". We omit these words in the figure and leave only the keywords.
We illustrate certain effects of our method in different columns.
The first two columns are about changing facial expression. Such experiments are very common in the GAN inversion methods~\cite{shen2020interpreting,xia2021survey}. Our method directly change the facial expressions without explicitly training a boundaries for attributes in advance like in~\cite{shen2020interpreting}.
The third column shows precise controllability of the age. In addition to \textit{young} and \textit{old}, more detailed instructions are also supported.
The images in the next four columns show results of different hairstyles, face shapes, makeups, and small changes around the eye area.
These results prove that our method can change the main structure of the face and precisely synthesize the tiny but nonnegligible attributes.
The visual attributes can be controlled with explicit descriptions, \eg, beard or blonde, as shown in the first three columns in Figure~\ref{fig:gen_open}. In the last column, we show some results implicitly controlled by indicating a celebrity, \ie, Joe Biden, Donald Trump, and Barack Obama.
`He looks like somebody' may refer to the identity or some important characteristics of these celebrities like hairstyle or skin color.
Obviously, our method regards these characteristics as more important than the identity. 
Figure~\ref{fig:with-clip-paint} demonstrates that our method can edit painting, caricature, and black and white photo images guided by textual descriptions without any re-training, fine-tuning, or post-processing.

\subsection{Region-of-interest Manipulation}
\label{subsec:roi}

This section illustrates some results of the region-of-interest editing introduced in Section~\ref{subsec:roi}.
We mask certain regions in the images and inpaint them with text guidance.
The results are shown in Figure~\ref{fig:man_roi}.
The first two rows are editing results of masked regions. The third row is for comparison. 
For the \texttt{single-word} descriptions \textit{blue} and \textit{red}, the results in the second row are primarily restricted to the designated regions.
For the third row, since the the editing areas are not designated, the editing maybe occurs in any possible regions, \eg, blue for eyes and red for clothes.

%% file: sections/discussion.tex
\section{Ablation Study and Discussion}
\label{sec:ablation}
At present, we follow the principle of minimization. Each module is the most simplified without any elaborately-designed architectures or tricks.
Removing any module, the output result cannot be obtained. For each module, we also use the most simple solutions, \eg, the L2-norm in visual-linguistic similarity learning in Section~\ref{subsec:vls-model} or the gradient-based algorithm  Adam~\cite{kingma2014adam} for optimization in Section~\ref{subsec:pretrained-text-encoder}. Adding well-designed modules can undoubtedly improve performance, but this goes against our original intention.
Thus, we organize the ablation study by illustrating the effectiveness of each module, discussing the weakness, and analyzing possible reasons.

\noindent\textbf{Instance-Level Optimization.}
The comparison of with or without instance-level optimization is shown in Figure~\ref{fig:inversion_result}. 
As shown in Figure~\ref{fig:inversion_result} (c), the inversion results of the image encoder preserve all attributes of the original images, which is sufficient for text-to-image generation since there is no identity to preserve. 
Manipulating a given image according to a text, however, should not change the unrelated attributes especially one's identity, which is preserved after the instance-level optimization (Figure~\ref{fig:inversion_result} (d)).  
We also compare with another strategy to preserve identity proposed by a recent inversion-based image synthesis method pSp~\cite{richardson2020encoding} that incorporates a dedicated recognition loss~\cite{deng2019arcface} during training.
Both methods can find a reasonable latent code for real face images.
Despite both achieving the goal of identity preservation, the optional instance-level optimization makes our framework concise and easy to train, and provides a non-deterministic way to refine the final results accordingly.
To be specific, pSp produces one deterministic result while ours provides a non-deterministic way that can refine the final results accordingly. 
For example, one can choose to directly output the results of the encoder, or optimize for steps. Through our empirical observation, 20 iterations are good enough to preserve identity.

\noindent\textbf{Visual-Linguistic Similarity.}
As AttnGAN shown, the global sentence vectors are used to ``sketch the primitive shape and colors of objects''.
This inspires us that the global information of facial images can be seen as main structures such as face shape, which is consistent with the hierarchically semantic property of Style-based generators.
The text encoder is trained using our visual-linguistic similarity and a very simple pairwise ranking loss~\cite{dong2017semantic} to align text and image embedding, as described in Section~\ref{subsec:vls-model}.
This pairwise ranking loss maximizes the cosine similarity of mismatching pairs, and minimizes the similarity of matched pairs.
Although the learned text embedding can handle near-miss cases, as shown in Figure~\ref{fig:near-miss}, we found this plain strategy sometimes may lead to insufficient disentanglement of attributes and mismatching of image-text alignment, leaving some room for improvement.

\figpaintclip
\figmanroi

\noindent\textbf{Potential Issue with StyleGAN.}
In our experiments, we found that some text-unrelated attributes are unwantedly changed when manipulating a given image according to a text description.
In the first place, we thought this may be caused by the proposed visual language similarity learning module. 
However, when performing layer-wise style-mixing on the inverted codes of two real images, it turns out that such interference still occurs.
This means some important facial attributes remain entangled in the $\mathcal{W}$ space. 
Under ideal circumstances, different facial attributes should be orthogonal (meaning without affecting other attributes).
Another inherent defect of StyleGAN is that some attributes, such as hats, necklaces, and earrings, are not well represented in its latent space. 
This sometimes makes our method perform less satisfactorily when adding or removing jewelry or accessories through natural language descriptions.

\section{Limitation and Outlook}
\label{sec:limitation}

\noindent\textbf{Expensive Image-specific Optimization.} 
Due the low-dimensionality of the latent code, the directly inverted results often do not resemble the given images. 
Typically, a slight difference between the identities of the manipulated result and the given image is observed.
To preserve identities and unrelated regions, we use iterative optimization during inference in both strategies of our method.
As such preservation usually require expensive image-specific optimization at runtime, the whole process is often time-consuming.
Some recent methods~\cite{wang2021gfpgan,chan2021glean,pan2020exploiting} of image restoration found that inversion-based methods often generate images with low-fidelity. 
Unlike most GAN inversion methods, which also use pre-trained GANs, they employ the pre-trained GANs as prior, \eg, designing a latent bank in a succinct encoder-bank-decoder architecture.
To get rid of restrictions, instead of directly using the unmodified latent space of the style-based generators, we can use these pre-trained GANs to provide rich and diverse priors. The textual guidance can be integrated in a similar way as in the previous text-to-image generation methods~\cite{xu2018attngan,li2019control}.

\noindent\textbf{One Certain Class Limitation.} 
It is a natural idea to adopt our method in dozens of classes where the StyleGAN generates high-resolution images, like cars, birds or any categories in the lsun dataset~\cite{yu2015lsun}.
Due to the limitation of StyleGAN itself, however, the state-of-the-art performance of our method is within a narrow image domain.
Style-based generators, not limited to StyleGAN, are suitable for images with certain classes and structures like faces~\cite{karras2019style,xia2021tedigan}, birds~\cite{wah2011caltech}, or bedrooms~\cite{yu2015lsun}. 
It may perform less-satisfactory on datasets that contain various objects or scenes like COCO~\cite{lin2014microsoft} or ImageNet~\cite{deng2009imagenet}. 
Some recent generation or restoration methods using pretrained GANs as prior are also limited to certain classes~\cite{wang2021gfpgan,chan2021glean}. 
Training a GAN to support the synthesis of images from many classes is still a challenging problem, especially when the fine-grained representations with multiple semantic levels are required. 
We notice that a concurrent zero-shot text-to-image system called DALL-E~\cite{ramesh2021zero} can generate arbitrary classes of images from open-world descriptions. It is trained on a far broader variety of unconstrained images and texts. 
Roughly speaking, DALL-E is a combination of three separately-trained parts: image encoder and decoder dVAE, text-encoder Transformer, and text-image matching CLIP.
Yet notwithstanding the breakthrough of the limitations of one certain class, the quality and resolution of its results are very limited.  
This limitation of DALL-E is primarily inherited from the dVAE. 
To break current limitations of our method on open-class and DALL-E on high-resolution, a novel generator that can generate images of arbitrary scenes or structures with plausible quality and resolution is indispensable.

\figinversionresult
\fignearmiss

\noindent\textbf{Supported Modalities.} 
In addition to text, label, and sketch that supported in our method, other modalities, like 3D shapes or sounds, can also be used as intermediaries in visual content creation. 
Is it possible for our proposed method to support 3D shapes or sounds? Or in other words, what if we want to generate and manipulate 2D images or 3D shapes through oral instructions directly without ``lifting a finger''?
The simplest way to support verbal orders, \ie, the modality of sound, although not elegant enough, is to turn the sound into the text input we need through a speech recognition system.
To generate 3D shapes directly from text or sound, we can easily integrate our method with a concurrent shape reconstruction method~\cite{pan2020gan2shape} for the reason that we share the same latent space of a pretrained GAN model.
In the future, we will exploit elegant solutions for efficient language-driven (including speech and word) generation and manipulation of 2D images and 3D shapes.

%% file: sections/conclusion.tex
\section{Conclusion}
\label{sec:conclusion}
In this work, we propose a novel method for image synthesis using textual descriptions, where two strategies are developed to use the strong generative powers of StyleGAN.
Both strategies can unify the two different tasks (text-guided image generation and manipulation) into the same framework and achieves high accessibility, diversity, controllability, and accurateness for facial image generation and manipulation.
Through the two proposed strategies using a pretrained StyleGAN as prior and a large-scale multi-modal face dataset, our method can effectively synthesize images from multi-modal inputs with unprecedented quality.
Extensive experimental results demonstrate the superiority of our method, in terms of the effectiveness of image synthesis, the capability of generating high-quality results, the extendability for multi-modal inputs, and the controllability of region-of-interest editing.